\documentclass[pdflatex,sn-basic]{sn-jnl}

\usepackage{graphicx}%
\usepackage{multirow}%
\usepackage{amsmath,amssymb,amsfonts}%
\usepackage{amsthm}%
\usepackage{mathrsfs}%
\usepackage[title]{appendix}%
\usepackage{xcolor}%
\usepackage{textcomp}%
\usepackage{manyfoot}%
\usepackage{booktabs}%
\usepackage{algorithm}%
\usepackage{algorithmicx}%
\usepackage{algpseudocode}%
\usepackage{listings}%

\usepackage{hyperref}

\usepackage{natbib}

\theoremstyle{thmstyleone}%
\theoremstyle{thmstyletwo}%

\theoremstyle{thmstylethree}%

\begin{document}

\title[Article Title]{Vector-Symbolic Architecture for Event-Based Optical Flow}


\author*[1]{\fnm{Hongzhi} \sur{You}}\email{hongzhi-you@uestc.edu.cn}

\author[2]{\fnm{Yijun} \sur{Cao}}\email{caoyijun@guet.edu.cn}

\author[1]{\fnm{Wei} \sur{Yuan}}\email{yw.zccz@gmail.com}

\author[1]{\fnm{Fanjun} \sur{Wang}}\email{fanjun.wang@std.uestc.edu.cn}


\author[3]{\fnm{Ning} \sur{Qiao}}\email{ning.qiao@synsense.ai}

\author[1]{\fnm{Yongjie} \sur{Li}}\email{liyj@uestc.edu.cn}

\affil*[1]{\orgdiv{MOE Key Laboratory for NeuroInformation}, \orgname{School of Life Science and Technology, University of Electronic Science and Technology of China}, \orgaddress{\city{Chengdu}, \state{Sichuan}, \country{China}}}

\affil[2]{\orgname{School of Electronic Engineering and Automation, Guilin University of Electronic Technology}, \orgaddress{\city{Guilin}, \state{Guangxi}, \country{China}}}

\affil[3]{\orgname{SynSense Tech. Co. Ltd.}, \orgaddress{\city{Ningbo}, \state{Zhejiang}, \country{China}}}


\abstract{From a perspective of feature matching, optical flow estimation for event cameras involves identifying event correspondences by comparing feature similarity across accompanying event frames. In this work, we introduces an effective and robust high-dimensional (HD) feature descriptor for event frames, utilizing Vector Symbolic Architectures (VSA). The topological similarity among neighboring variables within VSA contributes to the enhanced representation similarity of feature descriptors for flow-matching points, while its structured symbolic representation capacity facilitates feature fusion from both event polarities and multiple spatial scales. Based on this HD feature descriptor, we propose a novel feature matching framework for event-based optical flow, encompassing both model-based (VSA-Flow) and self-supervised learning (VSA-SM) methods. In VSA-Flow, accurate optical flow estimation validates the effectiveness of HD feature descriptors. In VSA-SM, a novel similarity maximization method based on the HD feature descriptor is proposed to learn optical flow in a self-supervised way from events alone, eliminating the need for auxiliary grayscale images. Evaluation results demonstrate that our VSA-based method achieves superior accuracy in comparison to both model-based and self-supervised learning methods on the DSEC benchmark, while remains competitive among both methods on the MVSEC benchmark. This contribution marks a significant advancement in event-based optical flow within the feature matching methodology.}

\keywords{Vector-symbolic architecture, Optical flow, Event camera, feature matching}



\maketitle

\section{Introduction}
\label{sec:intro}

Event-based cameras are bio-inspired vision sensors that asynchronously provide per-pixel brightness changes as an event stream \cite{ gallego2020event}. Leveraging their high temporal resolution, dynamic range, and low latency, these cameras have the potential to enhance accurate motion estimation, particularly in optical flow \cite{benosman2013event, almatrafi2020distance}. However, event-based optical flow estimation poses challenges due to its asynchronously and sparsely event visual information and the difficulty in obtaining ground-truth for optical flow, as compared to traditional cameras \cite{gallego2020event, shiba2022secrets}. Therefore, it is crucial to develop unsupervised optical flow methods that capitalize on the unique characteristics of event data, eliminating the dependency on expensive-to-collect and error-prone ground truth \cite{shiba2022secrets}.

Optical flow estimation involves finding pixel correspondences between images captured at different moments. The feature matching method, a fundamental approach for event-based optical flow, relies on maximizing feature similarity between accompanying frames \cite{gallego2020event}. In this method, the feature for each event is typically represented by the image pattern around the corresponding pixel in the event frame \cite{liu2018adaptive, liu2022edflow}. However, the inherent randomness in events \cite{gallego2020event} result in inconsistent image patterns of the same object across various frames, posing challenges in acquiring accurate and robust feature descriptors. Due to the absence of an effective event-only local feature descriptor, the feature matching method for event-based optical flow is generally limited to estimating sparse optical flows for key points, showing suboptimal performance \cite{liu2018adaptive, liu2022edflow}. During self-supervised learning, accurate dense optical flow estimation becomes challenging without restoring luminance or additional sensor information, such as grayscale images \cite{zhu2018ev, hagenaars2021self, deng2021learning, ding2022spatio}.

In this study, we introduce a high-dimensional (HD) feature descriptor for event frames, leveraging the Vector Symbolic Architecture (VSA). VSAs, regarded for their effectiveness in utilizing high-dimensional distributed vectors \cite{kleyko2021survey, kleyko2023survey}, have traditionally been employed in symbolic representations of artificial shapes \cite{karunaratne2021robust, renner2022neuromorphicVO, renner2022neuromorphic, hersche2023neuro} or few-shot learning classification tasks \cite{hersche2022constrained, karunaratne2022memory}. In this work, VSAs form the basis of our novel descriptor in natural scenes captured by event cameras. This descriptor utilizes the local similarity characteristics of neighboring variables within VSA \cite{frady2021computing, renner2022neuromorphicVO} to reduce the impact of randomness in events on representation accuracy. Employing structured symbolic representations \cite{komer2020biologically}, it achieves multi-spatial-scale and two-polarity feature fusion for feature descriptor. Our evaluation of descriptor similarity for flow-matching points on datasets DSEC and MVSEC demonstrates the effectiveness of our proposed approach. 

Further, we focus on a unifying framework for event-based optical flow within the feature matching strategy, centered around the proposed HD feature descriptors. The model-based VSA-Flow method, derived from the framework, utilizes the similarity of HD feature descriptors to achieve more accurate dense optical flow. Similarity integration in the cost volume from three event frame pairs with progressively doubling time intervals at gradually downsampled scales enables VSA-Flow to achieve large optical flow estimation within a limited neighboring region. Meanwhile, the proposed VSA-SM method relies on a similarity maximization (SM) proxy loss for predicted flow-matching points. This novel self-supervised learning approach effectively estimates optical flow from event-only HD feature descriptors, eliminating the need for additional sensor information. Evaluation results reveal that we obtain the best accuracy in both model-based and self-supervised learning methods on the DSEC-Flow benchmark, and competitive performance on the MVSEC benchmark.

\section{Related Works}
\subsection{Event-based Optical Flow Estimation}
\noindent From a methodological perspective, event-based optical flow estimation encompasses three primary approaches \cite{gallego2020event}. The first approach involves the gradient-based method, which leverages the spatial and temporal derivative information provided by event data directly or after appropriate processing to compute optical flow \cite{benosman2012asynchronous, benosman2013event}. Previous studies have explored event-based adaptations of Horn-Schunck and Lucas-kanade \cite{horn1981determining, lucas1981iterative, benosman2012asynchronous, almatrafi2019davis}, distance surface \cite{almatrafi2020distance, brebion2022real} and spatial-temporal plane-fitting \cite{benosman2013event, akolkar2020real}. 


The second approach is the feature matching method, which calculates optical flow by evaluating the similarity or correlation of feature representations for individual pixels between consecutive event frames in the temporal domain. For instance, the model-based EDFLOW estimates optical flow by applying adaptive block matching \cite{liu2018adaptive, liu2022edflow}. Meanwhile, this approach is frequently employed in the design of learning-based optical flow neural networks that incorporate cost volume modules capable of computing feature similarity or correlation, such as E-RAFT \cite{gehrig2021raft} and TMA \cite{ye2023towards}. In addition, treating auxiliary grayscale images as low-dimensional features, EV-FlowNet engages in self-supervised learning by minimizing the intensity difference between warped images based on the estimated optical flow \cite{zhu2018ev}. 

The third approach, exclusive to event cameras, is the contrast maximization method. This method maximizes an objective function, often related to contrast, to quantify the alignment of events generated by the same scene edge \cite{stoffregen2018simultaneous, gallego2018unifying, gallego2019focus}. The underlying idea is to estimate motion by reconstructing a clear motion-compensated image of the edge patterns that triggered the events. This approach can be applied not only to model-based optical flow estimation \cite{shiba2022secrets} but also frequently serves as a loss function for unsupervised and self-supervised optical flow learning \cite{shiba2022secrets, ye2020unsupervised, paredes2021back, hagenaars2021self,  paredes2023taming}.

In contrast to prior work, our proposed VSA-based framework for event-based optical flow adopts a classical feature matching approach to offer deeper insights into the problem. This framework is adaptable to both model-based and self-supervised learning methods, akin to the contrast maximization method \cite{gallego2018unifying, shiba2022secrets}. Particularly, the self-supervised learning method in the framework can achieve accurate optical flow solely from event-only VSA-based HD feature descriptors, eliminating the need for auxiliary grayscale images.

\subsection{High-dimensional Representations of Images Using Vector Symbolic Architecture}
\noindent Vector Symbolic Architectures (VSAs) are regarded as a powerful algorithmic framework that leverages high-dimensional distributed vectors and employs specific algebraic operations and structured symbolic representations \cite{kleyko2021survey, kleyko2023survey}. VSAs have demonstrated remarkable capabilities in various domains, including spatial cognition and visual scene understanding. The hypervector encoding of the color images and event frames, including artificial shapes, is achieved through a superposition of spatial index vectors, weighted by their corresponding image pixel values \cite{renner2022neuromorphic, renner2022neuromorphicVO}. These HD representations finds application in neuromorphic visual scene understanding \cite{renner2022neuromorphic} and visual odometry \cite{renner2022neuromorphicVO}. Leveraging the structured symbolic representation capacity of VSAs, a biologically inspired spatial representation is employed to generate hierarchical cognitive maps, each containing objects at various locations \cite{komer2020biologically}. Moreover, several VSA-based approaches have been introduced as frameworks for systematic aggregation of image descriptors suitable to visual place recognition \cite{neubert2021hyperdimensional, kempitiya2022parameterization}. Overall, VSA endows HD representations of images with intrinsic attributes of hierarchical structure and semantics. 

Accurate representations of feature descriptors that encompass individual pixels and their contextual features are crucial for optical flow estimation based on the feature matching method. In contrast to prior work, we adopt a specific type of VSA, Vector Function Architecture (VFA), which embodies continuous similarity characteristics to reduce the impact of randomness in events. This specific VSA is employed as a HD kernel to extract localized feature information from event frames. Meanwhile, optical flow estimation models commonly incorporate a multi-scale pyramid design to enhance their performance. Utilizing the binding capacity of structured features in VSA, we amalgamate HD feature representations from multiple scales and two event polarities into a unified feature descriptor.

\section{Methodology}

\subsection{Preliminary}

VSAs constitute a family of computational models with vector representations that have two distinct properties \cite{kleyko2021survey, frady2021computing}. Firstly, symbols are represented by mutually orthogonal randomized $d$-dimensional vectors ($\in \mathbb{R} ^d$), which facilitates a clear distinction between different symbols. Secondly, all computations within VSAs can be composed by a limited set of elementary vector algebraic operations, where the primary operations are the binding ($\circ$) and superposition ($+$) operations. The binding operation commonly signifies associations between symbols, such as a roll-filler pair \cite{kanerva2009hyperdimensional}, while the superposition operation is frequently used to represent sets of symbols. Both operations do not change the hypervector dimensionality. Through the combination of these operations and symbols, VSAs can effectively achieve structured symbolic representations. For instance, consider a scenario in which the character $\boldsymbol{1}$ is located at position $\boldsymbol{P_A}$ and $\boldsymbol{2}$ at position $\boldsymbol{P_B}$ in a given image. The hypervector symbolic representation of this image can be denoted as $\boldsymbol{I} = \boldsymbol{P_A} \circ \boldsymbol{One} + \boldsymbol{P_B} \circ \boldsymbol{Two}$, where $\boldsymbol{P_A}$, $\boldsymbol{P_B}$, $\boldsymbol{One}$ and $\boldsymbol{Two}$ $\in \mathbb{R} ^d$ represent mutually orthogonal randomized hypervectors of corresponding concepts. 

VSAs have various models that use different types of random vectors \cite{kleyko2021survey}. In this study, an improved Holographic Reduced Representation (HRR) is employed as the VSA model to ensure high concept retrieval efficacy \cite{ganesan2021learning}. For HRR, the binding operation is the circular convolution of two hypervectors, and the superposition operation the component-wise sum. Additionally, the similarity between two HRRs can be measured through the cosine similarity. 

In this work, the feature extraction from event frames requires the VSA-based 2-D spatial representation. Here, we first introduce the fractional power encoding (FPE) method \cite{plate1992holographic, plate1994distributed} for representing integers along each coordinate axis in an image plane, and then the VSA-based spatial representation.

\subsubsection{The Fractional Power Encoding Method}
In the fractional power encoding method \cite{plate1994distributed}, let $x\in \mathbb{Z}$ be an integer, $X\in \mathbb{R} ^d$ be a random hypervector, the hypervector representation $\mathbf{z}\left( x \right) \in \mathbb{R} ^d$ for any integer $x$ can be obtained by repeatedly binding the base vector $X$ with itself $x$ times as follows:
\begin{equation}
\mathbf{z}\left( x \right) :=X^x=\left( X \right) ^{\left( \circ x \right)} =\mathcal{F} ^{-1}\left\{ \mathcal{F} \left\{ X \right\} ^x \right\}
\label{EQU:FPE}
\end{equation}
where the rightmost equation denotes the fractional binding operation by expressing it in the complex domain \cite{komer2020biologically, frady2021computing}. $\mathcal{F} \left\{ \cdot \right\}$ is the Fourier transform, and $\mathcal{F} \left\{ \cdot \right\}^x$ is an component-wise exponentiation of the corresponding complex vector.

\subsubsection{The VSA-based Spatial Representation} 
Recent studies have demonstrated that the hypervector spatial representation $D\left( x,y \right)$ $\in \mathbb{R} ^d$ of a point ($x,y$) in 2-D space can be obtained using VSA with FPE \cite{komer2020biologically, frady2022computing}, as expressed in the following:
\begin{equation}
	D\left( x,y \right) =X^{x}\circ Y^y
	\label{EQU:SSP}
\end{equation}
Here, random vectors $X$ and $Y$ $\in \mathbb{R} ^d$ represent the base vectors for horizontal and vertical axes, respectively. $X^x$ and $X^y$ represent pseudo-orthogonal representation vectors for distinct integer positions $x$ and $y$ along each axes.

\subsection{The VSA-based Feature Matching Framework}

\noindent This work aims to establish a novel framework for event-based optical flow utilizing VSA, adaptable to both model-based and self-supervised learning methods within the feature matching approach. Optical flow estimation involves finding pixel correspondences between images captured at distinct time intervals. Effective event representation and precise feature descriptors are essential in the framework.

\subsubsection{Accumulative Time Surface}
\noindent Event cameras are innovative bio-inspired sensors that respond to changes in brightness through continuous streams of events $\mathcal{E} =\left\{ e_1,e_2,\cdots \right\}$ in a sparse and asynchronous manner. Each event $e_k=\left( x_k,y_k,t_k,p_k \right)$ comprises the space-time coordinates with polarity $p_k\in \left\{ +,- \right\}$. In this work, we use an event representation called accumulative Time Surface (TS) \cite{lagorce2016hots, zhou2021event}. An accumulative TS at pixel $(x,y)$ and time $t$ is defined as follows:
\begin{equation}
\mathcal{T}\left( x,y,t \right) \doteq \sum_{t_j\leq t}{\exp \left( -\frac{t-t_j\left( x,y \right)}{\tau _{TS}} \right)}
\label{EQU:TimeSurface}
\end{equation}
Here, $\tau _{TS}$ represents the exponential-decay rate, and $t_j$ denotes the timestamp of any event that occurred at pixel $(x,y)$ prior to time $t$. Thus, the accumulative TS emulates the synaptic activity that takes place after receiving the stream of events.

\subsubsection{VSA-based HD Kernel for Feature Extraction}

\begin{figure}[t]
\centering
\includegraphics[width=0.8\linewidth]{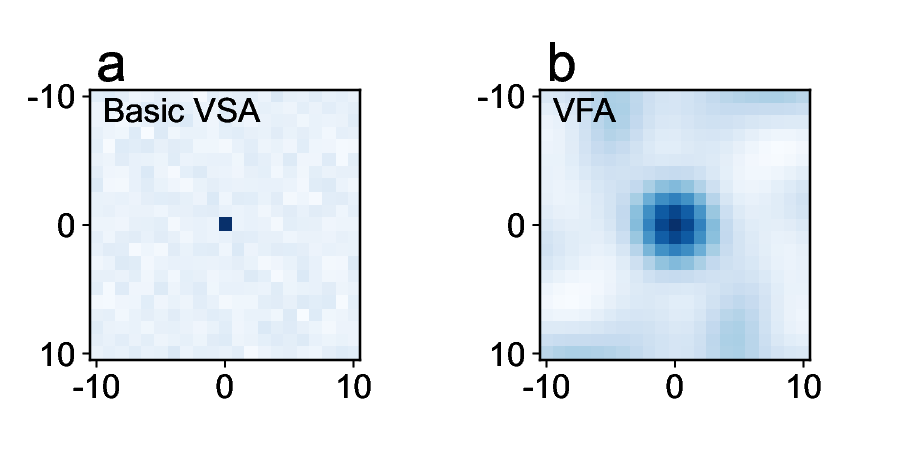} 
\caption{\textbf{Topological similarity in 2-D space for basic VSA and VFA HD kernels.} Similarity between hypervectors in the basic VSA (\textbf{a}) and VFA (\textbf{b}) HD kernels, respectively, originating from the center and surrounding points. The comparative analysis of hypervectors between the origin ($D(0,0)$) and points in its surrounding $N \times N$ neighborhood indicates that the VFA HD kernel, rather than the basic VSA HD kernel, is capable of capturing spatial topological similarity. $N=21$ ($n = \lfloor N/2 \rfloor = 10$).}

\label{fig:VSA_Kernel}
\end{figure}

Utilizing the spatial representation described in Equation \ref{EQU:SSP}, the HD feature representation $F\left( x,y \right) \in \mathbb{R}^{d}$ of the $N \times N$ neighborhood centered around the pixel ($x, y$) in the image $\mathcal{T} \in \mathbb{R}^{H \times W}$ can be encoded as a hypervector using the following formula \cite{renner2022neuromorphic}:
\begin{equation}
F\left( x,y \right) =\sum_{\varDelta x,\varDelta y}{\mathcal{T}\left( x+\varDelta x,y+\varDelta y \right) D\left( \varDelta x,\varDelta y \right)}
\label{EQU:HDRepresentation}
\end{equation}
where $(\varDelta x, \varDelta y)$ denotes the offset from the pixel $(x,y)$ to any pixel within its $N\times N$ neighborhood, in a scope of $\left[ -n,n \right]$, $n = \lfloor N/2 \rfloor$. From the perspective of 2-D image convolution, we can utilize $D$ $\in \mathbb{R}^{d \times N \times N}$ in Equation \ref{EQU:HDRepresentation} as the HD kernel to achieve local feature extraction within an $N \times N$ neighborhood for each pixel in the image. Consequently, the HD feature descriptor $F$ $\in \mathbb{R}^{d \times H \times W}$ of the image $\mathcal{T}$ can be efficiently obtained by convolving $\mathcal{T}$ with the HD kernel $D$ \cite{zhang1988shift} as follows:
\begin{equation}
F=\mathcal{T}  \ast D
\label{EQU:FeatureDescriptor}
\end{equation}

In principle, feature descriptors are required to capture differences between various image patterns of event frames, as well as exhibit similarities among comparable image patterns, displaying a certain degree of continuous similarity as image patterns vary. However, the basic VSA spatial representation defined in Equation \ref{EQU:SSP} and \ref{EQU:HDRepresentation} ignores important topological similarity relationships in 2-D space due to their pseudo-orthogonal property (Figure \ref{fig:VSA_Kernel}a) \cite{frady2022computing}. Given the inherent randomness in the event representations of the same object at different times, the spatial representation $D$ (Equation \ref{EQU:SSP}) is unsuitable as a HD kernel for feature extraction from event frames in tasks involving feature matching.

Recent studies have revealed that the Vector Function Architecture (VFA) \cite{frady2021computing} and the hyperdimensional transform \cite{dewulf2023hyperdimensional} exhibit continuous translation-invariant similarity kernels. Inspired from these findings and for the sake of simplicity, here we employ a Gaussian-smoothed HD kernel $K$ $\in \mathbb{R}^{d \times N \times N}$ with topological similarity to achieve the HD feature descriptors of the accumulative TS as follows:
\begin{equation}
	K=D*G
	\label{EQU:VFA}
\end{equation}
Here, $G$ represents a two-dimensional Gaussian kernel with a standard deviation of $\sigma_K$, facilitating the HD kernel $K$ to possess a translation-invariant similarity and characteristic similar to VFA (Figure 3, Equation 12 and Theorem 1 in \cite{frady2021computing}). Hence, we consider $K$ as specific instances of VFA. The corresponding hypervector spatial representation exhibits topological similarity relationships within a 2-D space (Figure \ref{fig:VSA_Kernel}b). Compared to the basic VSA (Equation \ref{EQU:SSP}), the local similarity characteristics of spatial representation in VFA (Equation \ref{EQU:VFA}) can effectively assist the feature descriptor in reducing the impact of randomness in events on representation accuracy. Unless explicitly noted, VSA used in the following sections is VFA.

\begin{figure*}[t]
\centering
\includegraphics[width=1\linewidth]{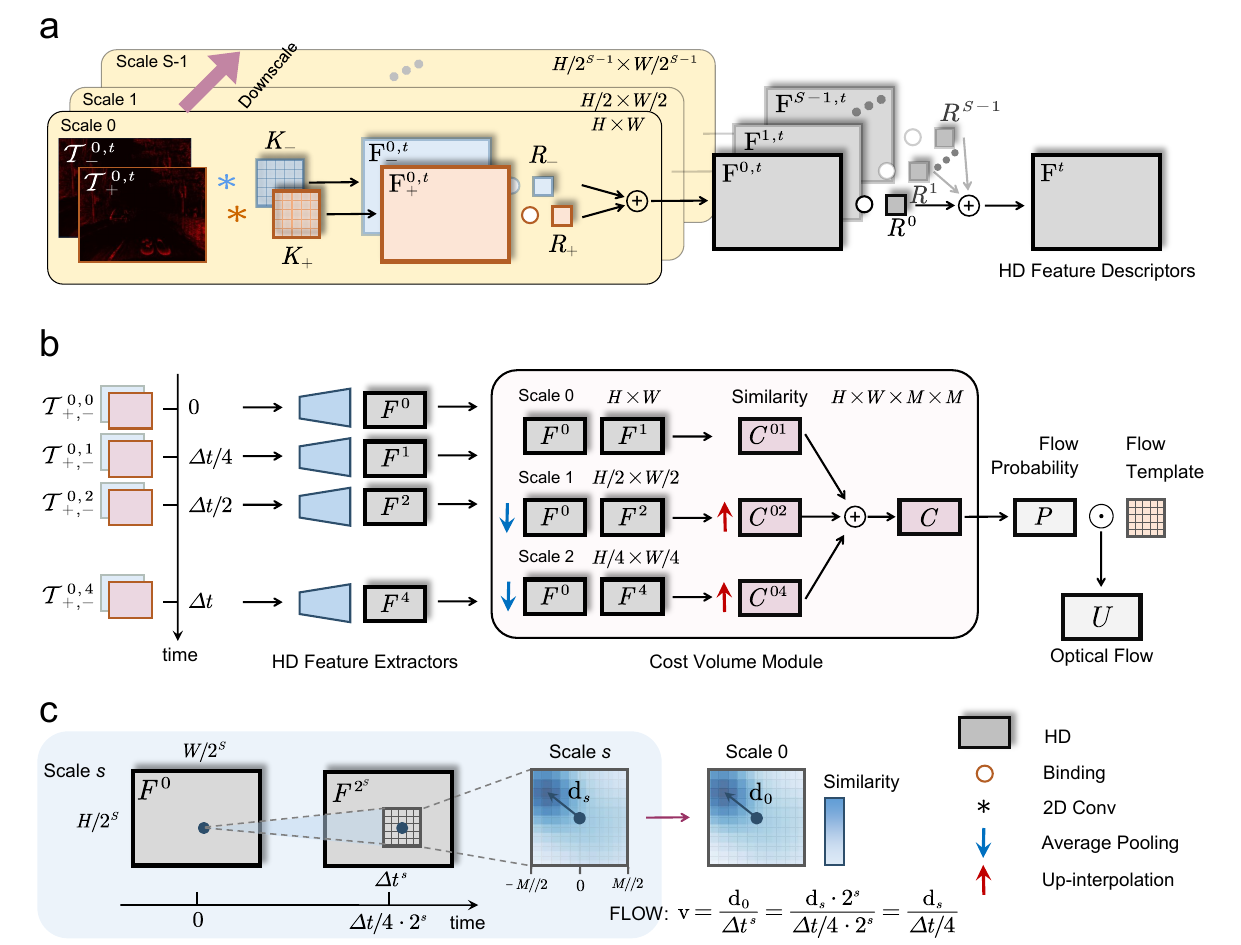} 
\caption{\textbf{Schematic of proposed VSA-Flow method for event-based optical flow.} 
	\textbf{(a)} Illustration of acquiring HD feature descriptors from accumulative TSs in a multi-scale strategy.
	\textbf{(b)} The VSA-Flow method consists of HD feature extractors, a cost volume module, and an optical flow estimator. HD feature extractors capture HD feature descriptors from TSs. The cost volume module computes local visual similarity by forming a volume representing similarity between $3$ TS pairs with different time intervals at different scales. The optical flow estimator generates flow using local visual similarity.
	\textbf{(c)} The mechanism allows for the direct fusion of three cost volumes at different scales through summation to form the final local visual similarity within the cost volume module.
}

\label{fig:Model_Architecture}
\end{figure*}

\subsubsection{VSA-based HD Feature Descriptor}


\noindent Inspired by classical estimation methods, feature descriptors at time $t$ are obtained by a multi-scale strategy \cite{black1996robust, memin2002hierarchical}. Here, the VSA-based HD feature descriptor involves three steps (Figure \ref{fig:Model_Architecture}a): transforming event streams into multiple scales of polarity-dependent accumulative TSs; generating HD feature descriptors for each scale by merging TSs from both polarities; and amalgamating HD feature descriptors from various scales into the final HD descriptor at the original scale of TSs. Here, we leverage the role-filler binding \cite{kleyko2021survey} to achieve the fusion of HD features, thereby realizing the structured representation of multi-scale and two-polarity HD feature descriptors. 

First, the accumulative TSs $\mathcal{T}_p\left(t \right) \in \mathbb{R}^{H \times W}$ for each polarity $p \in \{+,-\}$ at time $t$ are obtained from event streams according to Equation \ref{EQU:TimeSurface}. $\mathcal{T}_p\left(t \right)$ undergo continuous down-interpolation $S-1$ times at a ratio of $1/2$, resulting in a set of TSs $\mathcal{T}_p^{s,t} \in \mathbb{R}^{\frac{H}{2^s}\times \frac{W}{2^s}}$ ($s=0,1,\cdots S-1$). 

Second, utilizing the polarity-dependent HD kernel $K_p$ (Equation \ref{EQU:VFA}), the HD feature descriptor $F_{p}^{s,t}$ $\in \mathbb{R}^{d \times \frac{H}{2^s}\times \frac{W}{2^s}}$ for the corresponding TS of each polarity $p$ at Scale $s$ can be efficiently computed as follows:
\begin{equation}
F_{p}^{s,t}=\mathcal{T} _{p}^{s,t}*K_p
\label{EQU:F_mtp}
\end{equation}
By the role-filler binding, the HD feature descriptor $F^{s,t} \in \mathbb{R}^{d \times \frac{H}{2^s}\times \frac{W}{2^s}}$ for each scale is obtained from corresponding polarity-specific HD feature descriptors as follows: 
\begin{equation}
	F^{s,t}= F_{+}^{s,t}\circ R_++F_{-}^{s,t}\circ R_-
	\label{EQU:F_st}
\end{equation}
where $R_+$ and $R_-$ denote the random role (key) vectors for two polarities, respectively.

Finally, the HD feature descriptor $F^t$ $\in \mathbb{R}^{d \times H \times W}$ at time $t$, incorporating multiple spatial scales, can be represented as follows: 
\begin{equation}
F^t=\sum_{s=0}^{S-1}{F^{s,t} \circ R^s}
\label{EQU:F_t}
\end{equation}
Here, $R^s$ denote the random role (key) vectors for the corresponding spatial scale. In Equation \ref{EQU:F_t}, $F^{s,t}$ $\in \mathbb{R}^{d \times H \times W}$ is obtained through up-interpolation from $F^{s,t}$ in Equation \ref{EQU:F_st}.

\subsubsection{\textbf{Description of the Framework}}

Optical flow estimation involves identifying pixel correspondences between images captured at two different moments in time. The foundation of the feature matching method lies in the assumption that accurately estimated optical flow information corresponds to a higher similarity between corresponding pixels in accompanying event frames, compared to other pixels. The VSA-based feature matching framework here consists of two primary steps: 1) utilizing the VSA-based HD kernel to derive HD feature descriptors of consecutive event frames, and 2) employing algorithms such as search and optimization (for model-based methods) or neural networks with a proxy loss (for self-supervised learning methods). Both approaches aims to estimate optical flow by maximizing the similarity in feature descriptors of flow-matching points. In the following, we apply this framework to a model-based method (VSA-Flow) and a self-supervised learning method (VSA-SM) for event-based optical flow.

\subsection{VSA-Flow: A Model-based Method Using VSA}

\noindent The details of VSA-Flow is illustrated in Figure \ref{fig:Model_Architecture}b, comprising three main components: HD feature extractors, the cost volume module, and the flow generator. The HD feature extractors are responsible for obtaining corresponding VSA-based HD feature descriptors from the accumulative TSs essential for optical flow estimation. The cost volume module calculates local visual similarity by constructing a volume representing the similarity between all pairs of TSs. Finally, the optical flow estimator generates the optical flow based on the local visual similarity.

\subsubsection{\textbf{HD feature extractors}}

The accuracy of event-based optical flow estimation is hindered by the stochastic nature in events, especially when relying solely on two accumulative TS with a time difference of $\varDelta t$. To address this limitation and incorporate more comprehensive intermediate motion information into our method, we include accumulative TSs captured at time 0, $\varDelta t /4$, $\varDelta t/2$, and $\varDelta t$, each with two polarities, successively represented as $\mathcal{T}_{p}^{s,t}$ ($s=0$, $p \in \left\{ +,- \right\}$ and $t=0,1,2,4$) in Figure \ref{fig:Model_Architecture}b. By utilizing this extended set of event frames, we can achieve more precise optical flow estimation from time 0 to $\varDelta t$. Notably, the latter three time instances follow a progressive doubling pattern ($\times 2$), which will be further explained in the subsequent subsection. Following that, HD feature descriptors $F^t$ ($t=0,1,2,4$) corresponding to above event frames are acquired using the HD feature extractors depicted in Equation \ref{EQU:F_t} and Figure \ref{fig:Model_Architecture}a.

\subsubsection{\textbf{The cost volume module}}
Inspired by the basic cost volume in \cite{sun2018pwc, teed2020raft}, we adopt a strategy that integrates multiple pairs of HD feature descriptors with different time intervals: specifically, $F^0$ and $F^1$ at Scale $0$, $F^0$ and $F^2$ at Scale $1$, and $F^0$ and $F^4$ at Scale $2$ (Figure \ref{fig:Model_Architecture}b). The time interval $\varDelta t^s$ between $F^0$ and $F^{2^s}$ at Scale $s$ is $\varDelta t/4\cdot 2^s$ (Figure \ref{fig:Model_Architecture}c). The HD feature descriptors at the latter two scales are obtained through average pooling from those at Scale $0$ with kernel sizes $2$ and $4$, respectively, and equivalent stride. In this module, we first compute local visual similarity for each pair of HD descriptors $F^0$ and $F^{2^s} \in \mathbb{R}^{d \times H/2^s \times W/2^s}$ at Scale $s= 0,1,2$. Specifically, the HD descriptor of any event in $F^0$ is compared for similarity only with the descriptors of pixels within a surrounding $M \times M$ neighborhood in $F^{2^s}$ (Figure \ref{fig:Model_Architecture}c). Thus, the cost volume, $C^{02^s} \in \mathbb{R} ^{H/2^s\times W/2^s\times M\times M}$, can be efficiently computed using the cosine similarity as follows:
\begin{equation}
C_{ijkl}^{02^s}=\cos \left( F_{ij}^{0},F_{kl}^{2^s} \right) 
\label{EQU:CosSimilary}
\end{equation}

The displacement $\mathrm{d}_s$, which maps each event in $F^0$ to its corresponding coordinates in $F^{2^s}$, is obtained through the maximal similarity (Figure \ref{fig:Model_Architecture}c). The estimated optical flow $\mathrm{v}$ at the original scale of the event camera ($s=0$) can be calculated as follows:
\begin{equation}
\mathrm{v}=\frac{\mathrm{d}_0}{\varDelta t^s}=\frac{\mathrm{d}_s\cdot 2^s}{\varDelta t/4\cdot 2^s}=\frac{\mathrm{d}_s}{\varDelta t/4}\Longrightarrow \mathrm{d}_s=\frac{\varDelta t}{4}\mathrm{v}
\label{EQU:estimateflow1}
\end{equation}
Here, $\mathrm{d}_0$ denotes the corresponding displacement of $\mathrm{d}_s$ transformed from Scale $s$ to Scale $0$. Assuming the optical flow $\mathrm{v}$ is constant during the interval, Equation \ref{EQU:estimateflow1} reveals that $\mathrm{d}_s$ is independent of the scale and remains the same at different scales. This indicates that the cost volume, $C^{02^s}$ ($s=0,1,2$), should theoretically be the same for different scales. Hence, by up-interpolating all cost volumes at different scales to the same size at Scale 0, we obtain the final cost volume $C \in \mathbb{R} ^{H\times W\times M\times M}$ as the sum of all cost volumes (Figure \ref{fig:Model_Architecture}b). Here, similarity integration in the cost volume from three event frame pairs with progressively doubling time intervals at gradually downsampled scales enables VSA-Flow to achieve large optical flow estimation within a limited $M \times M$ neighboring region.

\subsubsection{\textbf{The optical flow estimator}} 
In this module, we adopt the scheme of optical flow probability volumes combined with a priori information on the position of the optical flow to estimate the optical flow (Figure \ref{fig:Model_Architecture}b) \cite{cao2023learning}. The optical flow probability volumes predict the probability of optical flow $P \in \mathbb{R} ^{H\times W\times M\times M}$ within a $M \times M$ local area for each pixel, based on the final cost volume $C$. The priori information on the position of the optical flow is provided by a predefined 2D grid template $T_{flow} \in \mathbb{R}^{2 \times M \times M}$ of optical flow containing all possible optical directions that align with the optical flow probability volumes. 

The optical flow probability volumes $P$ are calculated as follows:
\begin{equation}
\begin{cases}
	P_{ij}=\frac{\bar{C}_{ij}}{\sum_{k=1}^M{\sum_{l=1}^M{\bar{C}_{ijkl}}}}\\
	\bar{C}_{ij}=Max\left( C_{ij}-\alpha C_{ij}^{Max}-\left( 1-\alpha \right) C_{ij}^{Mean},0 \right)\\
\end{cases}
\label{EQU:FlowProbability}
\end{equation}
Here, $C_{ij}^{Max}$ and $C_{ij}^{Mean}$ represent the maximal and mean values of $C_{ij} \in \mathbb{R}^{M \times M}$. The coefficient $\alpha$ determines the probability area contributing to the estimated optical flow. The cost volume $C$ in Equation \ref{EQU:FlowProbability} is obtained by average pooling the original $C$ from the cost volume module with a kernel size of $s_c$ and a stride of $1$ to remove fluctuations due to the stochastic nature in events.

The template of optical flow $T_{flow}$ is formed by concatenating two 2D grids along the $x$ and $y$ axes with a range of $\left[ -m,m \right] \cdot 4/\varDelta t$, $m = \lfloor M/2 \rfloor$, respectively \cite{cao2023learning}:
\begin{equation}
T_x=\frac{4}{\varDelta t}\left[ \begin{matrix}
	-m&		\cdots&		\,\, m\\
	\vdots&		\ddots&		\vdots\\
	-m&		\cdots&		\,\, m\\
\end{matrix} \right] ,  T_y=\frac{4}{\varDelta t}\left[ \begin{matrix}
	-m&		\cdots&		-m\\
	\vdots&		\ddots&		\vdots\\
	\,\, m&		\cdots&		\,\, m\\
\end{matrix} \right] 
\label{EQU:FlowTemplate}
\end{equation}
Next, the optical flow $U$ $\in \mathbb{R}^{H\times W\times 2}$ is estimated from a weighted average of its probability volumes $P$ over the predefined template $T_{flow}$ (Figure \ref{fig:Model_Architecture}b) \cite{cao2023learning}, formulated as:
\begin{equation}
\begin{cases}
	U_x\left( i,j \right) =\sum_{k=1}^M{\sum_{l=1}^M{P_{ijkl}T_x\left( k,l \right)}}\\
	U_y\left( i,j \right) =\sum_{k=1}^M{\sum_{l=1}^M{P_{ijkl}T_y\left( k,l \right)}}\\
\end{cases}
\label{EQU:EstimatedFlow}
\end{equation}
where $U_x$ and $U_y$ represent the components of the predicted optical flow along the $x$ and $y$ axes, respectively.

\begin{figure}[t]
\centering
\includegraphics[width=0.6\linewidth]{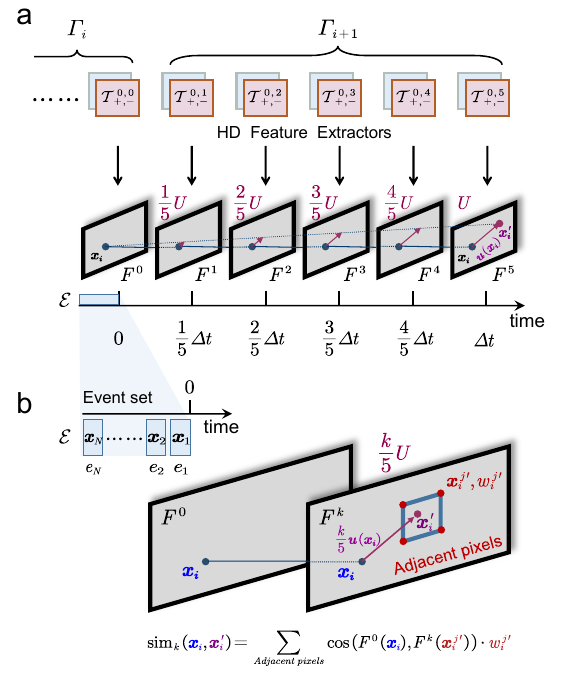} 
\caption{\textbf{Self-supervised optical flow learning via similarity maximization based on HD Feature descriptors.} 
	\textbf{(a)} Multi-frame approach for flow refinement. Within the time interval $\Delta t$, we utilize $K=5$ pairs of HD feature descriptors ($F^0\rightarrow F^k$,$k=1,...,K$) with progressively incremented intervals to compute the similarity between events and their corresponding matching points, ultimately enhancing the accuracy of optical flow estimation.
	\textbf{(b)} Illustration of similarity calculation for HD descriptors between events and their predicted flow-matching points. 
}

\label{fig:SelfLearning}
\end{figure}

\subsection{VSA-SM: A Self-supervised Learning Method Through Similarity Maximization}

\noindent Here, we adopt a self-supervised approach to learn optical flow estimation from accumulative TSs by maximizing the similarity of HD feature descriptors (Figure \ref{fig:SelfLearning}). We use a classical multi-frame approach for flow refinement, as illustrated in Figure \ref{fig:SelfLearning}a. During a time interval of $\Delta t$, we extract HD feature descriptors from corresponding accumulative TSs at intervals of $\Delta t/K$ ($K=5$), yielding a set of $K+1$ descriptors denoted as $F^k$ ($k = 0,...,K$). Assuming the optical flow within the interval $\Delta t$ is represented by $U$, the inferred optical flow between descriptor $F^0$ and descriptor $F^k$ equates to $kU/K$. As a result, we utilize $K$ pairs of descriptors ($F^0 \rightarrow F^k$, where $k=1,...,K$) to facilitate flow refinement within the context of self-supervised learning.

Knowing the per-pixel optical flow $\boldsymbol{u}\left( \boldsymbol{x} \right) \in U$, the matching point at time $\frac{k}{K}\varDelta t$ can be obtained through:
\begin{equation}
\boldsymbol{x}_{i}^{\prime}=\boldsymbol{x}_i+\frac{k}{K}\boldsymbol{u}\left( \boldsymbol{x}_i \right) 
\label{EQU:MatingPoints}
\end{equation}

However, the matching point $\boldsymbol{x}_{i}^{\prime}$ may not correspond to an actual pixel. Thus, the similarity between HD feature descriptors of $\boldsymbol{x}_i$ in $F^0$ and the matching point $\boldsymbol{x}_{i}^{\prime}$ in $F^k$ is calculated by evaluating its similarity to the descriptors of $4$ neighboring pixels ${\boldsymbol{x}_{i}^{j}}^{\prime}$ ($j=0,...,3$) around the matching point $\boldsymbol{x}_{i}^{\prime}$ in $F^k$, with normalized weights ${w_{i}^{j}}^{\prime}$ ($j=0,...,3$) via bilinear interpolation (Figure \ref{fig:SelfLearning}b):

\begin{equation}
\begin{cases}
	\mathrm{sim}_k\left( \boldsymbol{x}_i,\boldsymbol{x}_{i}^{\prime} \right) =\sum_{j}{\cos \left( F^0\left( \boldsymbol{x}_i \right) ,F^k\left( {\boldsymbol{x}_{i}^{j}}^{\prime} \right) \right) \cdot {w_{i}^{j}}^{\prime}}\\
	{w_{i}^{j}}^{\prime}=\frac{\kappa \left( x_{i}^{\prime}-{x_{i}^{j}}^{\prime} \right) \kappa \left( y_{i}^{\prime}-{y_{i}^{j}}^{\prime} \right)}{\sum_{j}{\kappa \left( x_{i}^{\prime}-{x_{i}^{j}}^{\prime} \right) \kappa \left( y_{i}^{\prime}-{y_{i}^{j}}^{\prime} \right) +\epsilon}}\\
	\kappa \left( a \right) =\max \left( 0,1-|a| \right), \epsilon \approx 0\\
\end{cases}
\label{EQU:Similarity}
\end{equation}

In this study, we use the similarity maximization proxy loss for feature matching to learn to estimate the event-based optical flow, as outlined in Equation \ref{EQU:Similarity}. Building upon the principles of previous unsupervised learning methods that emphasize contrast maximization \cite{shiba2022secrets, paredes2023taming}, we build on the loss function $\mathcal{L}$ as follows:
\begin{equation}
\mathcal{L} =\mathcal{L} _{similarity}+\lambda \mathcal{L} _{smooth}
\label{EQU:Total_Similarity}
\end{equation}
which is a weighted combination of two terms: the similarity loss $\mathcal{L} _{similarity}$ and the smoothness $\mathcal{L} _{smooth}$. The computation of the similarity loss involves $N$ pixels, encompassing the most recent events occurring before time $0$, as well as pixels sampled at every 5th interval both horizontally and vertically across the image plane (Figure \ref{fig:SelfLearning}a and \ref{fig:SelfLearning}b). The formulation of the similarity loss is as follows:
\begin{equation}
\begin{cases}
	\mathcal{L} _{similarity}=1-\left< \mathrm{sim} \right> ^{\alpha}\\
	\left< \mathrm{sim} \right> =\frac{1}{KN}\sum_{k=1}^K{\sum_{i=1}^N{\mathrm{sim}_k\left( \boldsymbol{x}_i,\boldsymbol{x}_{i}^{\prime} \right)}}\\
\end{cases}
\label{EQU:Loss_Similarity}
\end{equation}
Here, $\left< \mathrm{sim} \right>$ represents the average similarity encompassing all relevant pixels within $K$ pairs of descriptors, while $\alpha$ serves as a coefficient. A higher value of $\left< \mathrm{sim} \right>$ corresponds to more accurate optical flow estimation and a diminished similarity loss function. Additionally, the smoothness $\mathcal{L}_{\text{smooth}}$ adopts the Charbonnier smoothness prior \cite{ hagenaars2021self, zhu2019unsupervised} or the first order edge-aware smoothness \cite{stone2021smurf}.

In this study, we train E-RAFT \cite{gehrig2021raft} in a self-supervised manner, utilizing the loss function described in Equation \ref{EQU:Total_Similarity}, to demonstrate the effectiveness of our self-supervised learning method based on similarity maximization of HD feature descriptors. In principle, this methodology holds applicability across various event-based optical flow networks. Meanwhile, we adopt the full-image warping technique \cite{stone2021smurf} to improve flow quality near image boundaries.

\section{Experiments}

\subsection{Datasets, Metrics and Implementation Details}

Following previous works \cite{gehrig2021raft, shiba2022secrets},
both VSA-Flow and VSA-SM are evaluated using well-established event-based datasets DSEC-Flow ($640\times480$ pixel resolution) \cite{gehrig2021raft} and MVSEC ($346\times260$ pixel resolution) \cite{zhu2018ev}. 

For the model-based method (VSA-Flow), experiments are conducted on the official testing set of the public DSEC-Flow benchmark and on outdoor\_day1 and three indoor\_flying sequences with time intervals of $dt=1,4$ gray images on the MVSEC benchmark. For the self-supervised learning method (VSA-SM), E-RAFT is trained on the official training set of DSEC and on outdoor\_day2 sequence on MVSEC, respectively. To increase the variation in the optical flow magnitude during training, the training sequences on MVSEC are extended with time intervals of $dt=0.5,1,2,4,8$ gray images. Following separate training, evaluations are performed on the same testing sets as VSA-Flow, respectively on DSEC and MVSEC. Both methods are implemented using Pytorch library. For the VSA-SM training, we set batch size to $1$, the optimizer is set to Adam \cite{kingma2014adam} and learning rate to $1e{-2}$. 

We evaluate the accuracy of our predictions using following metrics: 
(i) EPE, the endpoint error; 
(ii) $\%_{1\mathrm{PE}}$ and $\%_{3\mathrm{PE}}$, 
the percentage of points with EPE greater than $1$ and $3$ pixels; 
(iii) AE, angular error. 
For both DSEC-Flow \cite{gehrig2021dsec, gehrig2021raft} and MVSEC \cite{zhu2018ev} datasets, 
metrics are measured over pixels with valid ground-truth and at least one event in the evaluation intervals.

\begin{figure*}[t]
	\centering
	\includegraphics[width=0.8\linewidth]{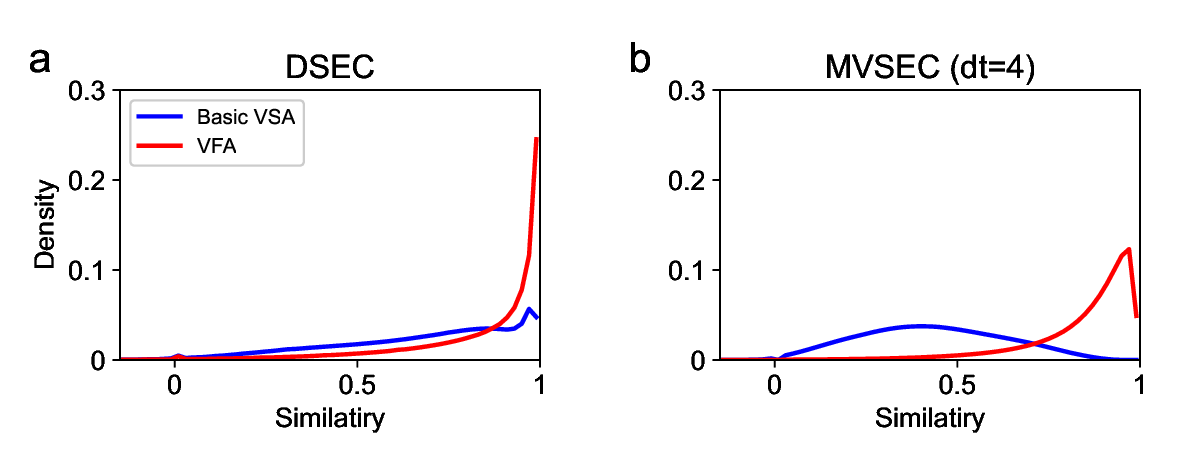} 
	\caption{\textbf{The probability density of similarity between matching points based on ground-truth (GT) optical flow on two datasets.} For each dataset, we can compute the similarity of HD feature descriptors for $N_{match}$ pairs of flow-matching points according to GT. The probability density of similarity refers to the likelihood that the feature similarity of flow-matching points equals a certain value. Compared to the basic VSA, VFA demonstrates enhanced capability in encoding the similarity of matching points in event frames.
	}
	
	\label{fig:DescriptorSIM}
\end{figure*}

\subsection{Descriptor Similarity of Flow-Matching Points}

In this study, HD feature descriptors are derived from feature extractors utilizing VSA-based HD kernels. We explore the impact of different VSA types (basic VSA and VFA) on the descriptor similarity among flow-matching points within the DSEC and MVSEC datasets (Figure \ref{fig:DescriptorSIM}). 

In the basic VSA HD kernel, all hypervectors are pseudo-orthogonal, implying that each pixel within the neighborhoods contributes independently to the feature descriptor. Feature descriptors obtained from the basic VSA HD kernel reflect the most fundamental image patterns. Hence, Figure \ref{fig:DescriptorSIM} (blue curves) reveals that the similarity of flow-matching points in the MVSEC dataset is inferior to that in the DSEC dataset. This observation suggests that, in comparison to the DSEC dataset, the MVSEC dataset experiences greater randomness in event frames, leading to lower event frame quality.

Figure \ref{fig:DescriptorSIM} (red curves) illustrates that VFA yields higher descriptor similarity for flow-matching points compared to basic VSA. In contrast to basic VSA, VFA exhibits an improved ability to encode the similarity of flow-matching points in event frames. 

\begin{table*}[htpb]
	\caption{Results on DSEC-Flow dataset \cite{gehrig2021raft}. Model-based (MB) methods need no training data; supervised learning (SL) methods need ground-truth; self-supervised learning (SSL) methods shown in this table only require events. Best accuracy presented in bold, and best accuracy in each type of method underlined. $^*$EV-FlowNet \cite{zhu2019unsupervised} is retrained by the corresponding literature. $^{\dagger}$$d=1024$, $N=21$ (the kernel size of HD feature descriptors), $S=2$; $K=5$. }
	\setlength{\tabcolsep}{2mm}
	\scriptsize
	\label{ResultsDSEC}
	\centering
	\begin{tabular}{llllll}
		\toprule
		& Methods & EPE & 1PE & 3PE & AE \\
		\midrule
		
		\multicolumn{1}{l}{\multirow{4}{3.3mm}{MB}}
		
		& MultiCM \cite{shiba2022secrets} &  3.47&  76.57 &  30.86 &  13.98 \\
		& RTEF \cite{brebion2022real} &  4.88&  82.81 &  41.96 &  10.82 \\
		\cmidrule{3-6}
		& \multicolumn{1}{l}{\multirow{1}{*}{VSA-Flow (VFA)$^{\dagger}$}} &  \underline{3.46$\pm$0.06}&  \underline{68.94$\pm$0.57} &  \underline{28.97$\pm$0.40} &  \underline{9.45$\pm$0.17} \\
		
		\cmidrule{3-6}
		& \multicolumn{1}{l}{\multirow{1}{*}{VSA-Flow (Basic VSA)$^{\dagger}$}} &  4.19$\pm$0.15 &  77.50$\pm$1.08 &  32.34$\pm$0.72 &  13.41$\pm$0.56 \\
		
		\midrule
		
		\multicolumn{1}{l}{\multirow{5}{1mm}{SL}}
		& EV-FlowNet \cite{gehrig2021raft}$^*$ &  2.32&  55.4 &  18.6 &  - \\
		& E-RAFT \cite{gehrig2021raft} &  0.79&  12.74 &  2.68 &  2.85 \\
		& IDNet \cite{wu2022lightweight} &  \underline{\textbf{0.72}}&  \underline{\textbf{10.07}} &  \underline{\textbf{2.04}} &  2.72 \\
		& TMA \cite{liu2023tma} &  0.74&  10.86 &  2.30 &  2.68 \\
		& E-Flowformer \cite{li2023blinkflow} &  0.76&  11.23 &  2.45 &  \underline{\textbf{2.68}} \\
		
		\midrule
		
		\multicolumn{1}{l}{\multirow{3}{1mm}{SSL}}
		& EV-FlowNet \cite{paredes2023taming}$^*$ &  3.86 &  - &  31.45 &  - \\
		& TamingCM \cite{paredes2023taming} &  2.33 &  68.29 &  17.77 &  10.56 \\
		& VSA-SM (VFA)$^{\dagger}$ &  \underline{2.22} &  \underline{55.46} &  \underline{16.83} &  \underline{8.86} \\
		

		\bottomrule
	\end{tabular}
	
	
\end{table*}

\begin{table*}[htbp]
	\centering
	\caption{Results on MVSEC dataset \cite{zhu2018ev}. SSL$_\mathrm{F}$: semi-supervised learning methods use grayscale images for supervision; Best accuracy presented in bold, and best accuracy in MB and SSL methods underlined..  $^{\dagger}$ Only used Scale 0 ($F^0\rightarrow F^4$) in the cost volume module and $K=1$ due to the small $\Delta t$, $d=1024$, $N=25$ and $S=2$; $^*$$d=1024$, $N=25$ and $S=2$. }
	\scriptsize
	\setlength{\tabcolsep}{0.7mm}
	\label{ResultsMVSEC1}
	\begin{tabular}{llcccccccc}
		\toprule
		\multirow{2}{2mm}{} & \multicolumn{1}{l}{\multirow{2}{20mm}{Methods}} & \multicolumn{2}{c}{indoor\_flying1} & \multicolumn{2}{c}{indoor\_flying2} & \multicolumn{2}{c}{indoor\_flying3} & \multicolumn{2}{c}{outdoor\_day1} \\
		\cmidrule{3-10}          & \multicolumn{1}{l}{} & EPE   & 3PE   & EPE   & 3PE   & EPE   & 3PE   & EPE   & 3PE \\
		
		\midrule
		\multicolumn{1}{l}{} & \multicolumn{9}{l}{$dt=1$}  \\
		\midrule

		\multicolumn{1}{l}{\multirow{5}{*}{MB}} & Nagata et al. \cite{nagata2021optical} & 0.62 & --- & 0.93 & --- & 0.84 & --- & 0.77 & --- \\
		
		& Akolkar et al. \cite{akolkar2020real} & 1.52 & --- & 1.59 & --- & 1.89 & --- & 2.75 & --- \\
		
		& Brebion et al. \cite{brebion2022real} & 0.52   & 0.10   & 0.98   & 5.50   & 0.71   & 2.10   & 0.53  & 0.20 \\
		
		&MultiCM \cite{shiba2022secrets} & \underline{\textbf{0.42}}  & 0.10   & \underline{\textbf{0.60}}   & \underline{\textbf{0.59}}   & \underline{\textbf{0.50}}  & \underline{\textbf{0.28}}   & \underline{0.30}   & \underline{0.10}  \\
		
		
		& VSA-Flow (VFA) $^{\dagger}$ & 0.46  & \underline{0.05}  & 0.65 & 1.08  & 0.53  & 0.29  & 0.65 & 3.60  \\
		\midrule
		
		\multicolumn{1}{l}{\multirow{3}{*}{SL}} & EV-FlowNet+ \cite{stoffregen2020reducing} & 0.56   & 1.00  & 0.66  & 1.00 & 0.59   & 1.00   & 0.68   & 0.99  \\
		
		& E-RAFT \cite{gehrig2021raft} & 1.10   & 5.72  & 1.94   & 30.79   & 1.66   & 25.20& \textbf{0.24}  & \textbf{0.00}  \\
		& TMA \cite{liu2023tma} & 1.06   & 3.63   & 1.81   & 27.29   & 1.58  & 23.26   & 0.25   & 0.07 \\
		
		\midrule

		\multicolumn{1}{l}{\multirow{3}{*}{SSL$_\mathrm{F}$}} & EV-FlowNet \cite{zhu2018ev} & 1.03  & 2.20  & 1.72   & 15.10   & 1.53   & 11.90  & 0.49   & 0.20  \\
		
		& Spike-FlowNet \cite{lee2020spike} & 0.84   & -----   & 1.28   & -----   & 1.11 & -----  & 0.49   & -----  \\
		& STE-FlowNet \cite{ding2022spatio} & 0.57   & 0.10   & 0.79  & 1.60   & 0.72  & 1.30   & 0.42   & \textbf{0.00}  \\
		
		\midrule
		
		\multicolumn{1}{l}{\multirow{3}{*}{SSL}} & EV-FlowNet \cite{stoffregen2020reducing} & 0.58  & \underline{\textbf{0.00}}   & 1.02  & 4.00   & 0.87   & 3.00   & \underline{0.32}   & \underline{\textbf{0.00}} \\
		& Hagenaars et al.\cite{hagenaars2021self} & 0.60  & 0.51  & 1.17   & 8.06   & 0.93   & 5.64 & 0.47  & 0.25  \\
		& VSA-SM (VFA)$^*$ & \underline{0.57}  & 0.07  & \underline{0.91}  & \underline{3.91}  & \underline{0.69}  & \underline{1.63}  & 0.46  & 3.42 \\
		
		
		\midrule
		\multicolumn{1}{l}{} & \multicolumn{9}{l}{$dt=4$}  \\
		\midrule
		
		\multicolumn{1}{l}{\multirow{2}{*}{MB}} &MultiCM \cite{shiba2022secrets} &  1.69  & 12.95  &   \textbf{2.49}  &   26.35  &   2.06  &   19.03  &   \underline{1.25}  &   \underline{9.21} \\
		
		
		
		& VSA-Flow (VFA) $^{\dagger}$ &  \underline{\textbf{1.44}} & \underline{\textbf{6.71}} &  \underline{\textbf{2.49}} &  \underline{\textbf{18.01}} & \underline{\textbf{1.79}} &  \underline{\textbf{11.90}} &  1.66 &  13.96 \\
		\midrule
		
		\multicolumn{1}{l}{\multirow{2}{*}{SL}} & E-RAFT \cite{gehrig2021raft} &  2.81  &  40.25  &  5.09  &  64.19  &  4.46  &  57.11  & 0.72  &  1.12 \\
		& TMA \cite{liu2023tma} &  2.43  &  29.91  &  4.32  &  52.74  &  3.60  &  42.02  &  \textbf{0.70}  &  \textbf{1.08} \\
		
		\midrule

		\multicolumn{1}{l}{\multirow{3}{*}{SSL$_\mathrm{F}$}} & EV-FlowNet \cite{zhu2018ev} &  2.25  &  24.70  &  4.05  &  45.30  &  3.45  &  39.70  &  1.23  &  7.30 \\
		
		& Spike-FlowNet \cite{lee2020spike} &  2.24  & -----  &  3.83  &  -----  &  3.18  &  -----  &  1.09  &  ----- \\
		& STE-FlowNet \cite{ding2022spatio} &1.77  &  14.70  &  2.52  &  26.10  &  2.23  &  22.10  &  0.99  &  3.90 \\
		
		\midrule
		
		\multicolumn{1}{l}{\multirow{3}{*}{SSL}} & EV-FlowNet \cite{stoffregen2020reducing} &  2.18  &  24.20  &  3.85  &  46.80  &3.18  & 47.80  &  1.30  &  9.70 \\
		& Hagenaars et al.\cite{hagenaars2021self} &  2.16  &  21.51  &  3.90  &  40.72  &  3.00  & 29.60  &  1.69  &  12.50 \\
		& VSA-SM (VFA)$^*$ &  \underline{1.63} &  \underline{10.05} &  \underline{2.92} &  \underline{22.57} & \underline{1.98} &  \underline{13.12} &  \underline{1.24} &  \underline{8.31} \\ 
		
		\bottomrule
		
	\end{tabular}%

\end{table*}%

\subsection{Results on DSEC}

Table \ref{ResultsDSEC} presents the evaluation results on the DSEC-Flow benchmark \cite{gehrig2021raft}. 
The methods listed in different rows are classified into three types: 
model-based methods (MB), supervised learning methods (SL), and self-supervised learning methods (SSL). 
The notations `VFA' and `Basic VSA' within the parentheses for our methods
represent the utilization of the VFA (Eq. \ref{EQU:VFA}) and the basic VSA (Eq. \ref{EQU:SSP}) HD kernels for feature descriptors.
It's important to note that the stochastic nature of generating spatial base vectors 
for the HD kernel impacts the evaluation of the VSA-Flow method, 
all evaluation metrics for the VSA-Flow method represent the statistical outcomes obtained 
from randomly producing 10 sets of HD kernels. 
This includes the mean and standard deviation for each metric. 
Regarding the VSA-SM method, due to its prolonged training duration, 
Table \ref{ResultsDSEC} showcases the evaluation results 
based on a single set of randomly generated HD kernel used during training.

The VSA-Flow (VFA) method provides the superior performance among all model-based methods 
in the DSEC-Flow dataset. 
In particular, the EPE and 3PE metrics slightly outperform other methods, 
whereas the 1PE and AE metrics display substantial improvements. Moreover, it is evident that employing VFA as the HD kernel in VSA-Flow leads to a significant performance improvement compared to utilizing the basic VSA, which is consistent with observations in Figure \ref{fig:DescriptorSIM}. In self-supervised training group, the proposed VSA-SM (VFA) method demonstrates the best results among all self-supervised learning methods. 
The extent of its improvement across the metrics aligns with the evaluation outcomes of the VSA-Flow (VFA).

\subsection{Results on MVSEC}

Table \ref{ResultsMVSEC1} reports the 
evaluation results on the MVSEC benchmark \cite{zhu2018ev}. Due to the small deviation of all metrics when $d=1024$ (Table \ref{ImpactDandSonDSEC}), for the sake of simplicity, the evaluation results on MVSEC for our methods come from a single set of randomly generated HD kernel. Consistent with \cite{zhu2018ev} and \cite{shiba2022secrets}, Table \ref{ResultsMVSEC1} compares some primary methods using the same training and testing sequences. Many learning-based methods trained on alternate outdoor sequences or datasets are not used for testing. 

The VSA-Flow method achieves the best results among all methods in indoor\_flying sequences when $dt=4$ and the competitive results when $dt=1$. These results indicate that the model-based VSA-Flow method, based on HD feature descriptors, are well-suited for large optical flow estimation ($dt=4$) and maintain competitiveness for low optical flow ($dt=1$). In addition, in comparison to the indoor\_flying sequences, the performance of VSA-Flow is less competitive in the outdoor\_day sequence.  This discrepancy may primarily stem from the fact that, compared to the indoor\_flying scene, the smaller motion in the outdoor\_day scene leads to sparser events \cite{zhu2019unsupervised}, thereby impacting the representation of HD feature descriptors. 

As mentioned earlier, the training sequences for VSA-SM on MVSEC are extended with time intervals of $dt=0.5, 1, 2, 4, 8$ grayscale images. Because VSA-Flow exhibits relatively weaker performance for low optical flow ($dt=1$, Table \ref{ImpactDandSonDSEC}) compared to large optical flow ($dt=4$, Table \ref{ImpactDandSonDSEC}), in the training strategy for VSA-SM, the optical flow predictions at time intervals $dt=0.5, 1, 2$ are scaled by factors of $8$, $4$, and $2$, respectively. Subsequently, self-supervised learning when $dt=0.5, 1, 2$ is conducted using high-dimensional feature descriptors of events frames for $dt=4$. Evaluation results indicate that the VSA-SM method achieves competitive performance compared to other self-supervised learning methods. Furthermore, it outperforms some semi-supervised learning methods that employ grayscale images for supervision, particularly on certain sequences.

It is noteworthy that many learning methods, including VSA-SM, exhibit lower performance in the indoor scenes compared to model-based methods. This discrepancy arises because training for MVSEC is exclusively conducted on the outdoor\_day2 sequence, but indoor and outdoor sequences contain distinct scene information.

 \begin{figure*}[t]
	\centering
	\includegraphics[width=1\linewidth]{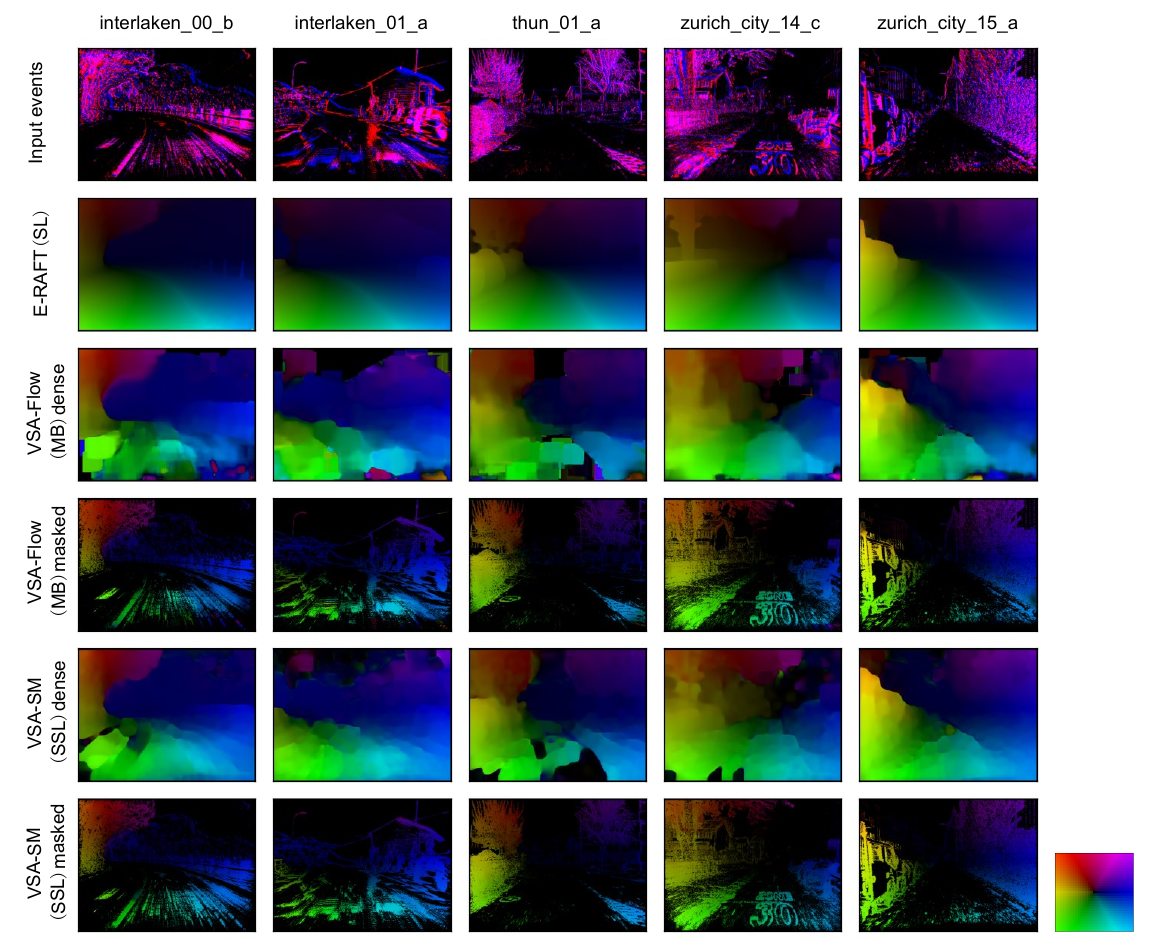} 
	\caption{\textbf{Qualitative comparision of our methods with the state-of-the-art E-RAFT architecture on several test sequence partitions of the DSEC dataset} \cite{gehrig2021raft}.
	}
	
	\label{fig:ResultsSampleDSEC}
\end{figure*}

\subsection{Qualitative Results on DSEC}

Qualitative results of both VSA-Flow and VSA-SM methods on multiple sequences from the test partition of DSEC-Flow dataset are shown in Figure \ref{fig:ResultsSampleDSEC}. Given the unavailability of ground truth for the official testing set, a comparison with the state-of-the-art E-RAFT architecture \cite{gehrig2021raft} is performed. Our model-based and self-supervised learning methods can achieve high-quality event-based optical flow estimation from events without the need for additional sensory information. Several conclusions can be drawn from these results: (1) Both VSA-Flow and VSA-SM accurately estimate optical flow, particularly in regions containing events. Event-masked sparse optical flow estimation appears more accurate than dense flow estimation. (2) The optical flow estimation from VSA-SM appears smoother compared to VSA-Flow; (3) VSA-Flow exhibits inaccuracies in optical flow estimation near image boundaries, whereas the adoption of a full-image warping technique \cite{stone2021smurf} for VSA-SM during self-supervised learning enhances its accuracy near image boundaries; (4) Due to both methods relying solely on event frames for flow estimation, accuracy diminishes in large areas devoid of events, sometimes resulting in zero flow estimation - a trend consistent with other self-supervised learning methods \cite{hagenaars2021self,paredes2023taming}; (5) As model-based and self-supervised learning approaches relying on event-only local features, our methods predict optical flow less smoothly compared to supervised learning methods; Meanwhile, our methods exhibit less sharp optical flow estimation at the edges of objects, displaying a smoother transition. 

\begin{table*}[ht]
	\caption{Impact of $d$ and $S$ for VSA-Flow (VFA) Method on DSEC-Flow dataset \cite{gehrig2021raft}. $d$: the hypervector dimension; and $S$: the multi-scale number in the VSA-based HD feature descriptor. }
	\setlength{\tabcolsep}{2mm}
	\scriptsize
	\label{ImpactDandSonDSEC}
	\centering
	\begin{tabular}{llllll}
		\toprule
		\makebox[0.08\textwidth][l]{$d$} & $S$ & EPE & 1PE & 3PE & AE \\
		\midrule
		
		\multicolumn{1}{l}{\multirow{3}{1mm}{1024}}
		
		& 1 &  \textbf{3.40}$\pm$0.05 &  70.26$\pm$1.17 &  \textbf{28.36}$\pm$0.49 &  9.93$\pm$0.31 \\
		& 2 &  3.46$\pm$0.06 &  \textbf{68.94}$\pm$0.57 &  28.97$\pm$0.40 &  \textbf{9.45}$\pm$0.17 \\
		& 3 &  3.85$\pm$0.06 &  69.81$\pm$0.46 &  31.35$\pm$0.28 &  9.64$\pm$0.10 \\
		
		\midrule

		512 &   &  3.56$\pm$0.07 &  69.44$\pm$1.20 &  29.55$\pm$0.56 &  9.73$\pm$0.32 \\
		256 & 2 &  3.63$\pm$0.12 &  70.10$\pm$1.25 &  29.97$\pm$0.84 &  10.00$\pm$0.43 \\
		128 &   &  3.87$\pm$0.28 &  72.71$\pm$2.21 &  31.75$\pm$1.64 &  11.08$\pm$1.02 \\

		\bottomrule
	\end{tabular}
	
	
\end{table*}

\subsection{Effects of Hypervector Dimension and Multi-scale}
Table \ref{ImpactDandSonDSEC} reports evaluation results for  experiments on the VSA-Flow method 
with varying hypervector dimensions ($d$) and different multi-scale numbers ($S$) 
for the HD feature descriptor. When $d=1024$, the VSA-Flow exhibits better EPE and 3PE metrics with $S=1$, while for $S=2$, 
it demonstrates better 1PE and AE metrics. 
Moreover, as $S$ remains constant and $d$ is altered, 
all metrics are improved with $d$, indicating that a larger hypervector dimension leads to enhanced performance. 
This result is consistent with the understanding that, within VSA, 
increased hypervector dimensions contribute to heightened information encoding capabilities \cite{kleyko2021survey, kleyko2023survey}.

\begin{figure}[t]
	\centering
	\includegraphics[width=0.7\linewidth]{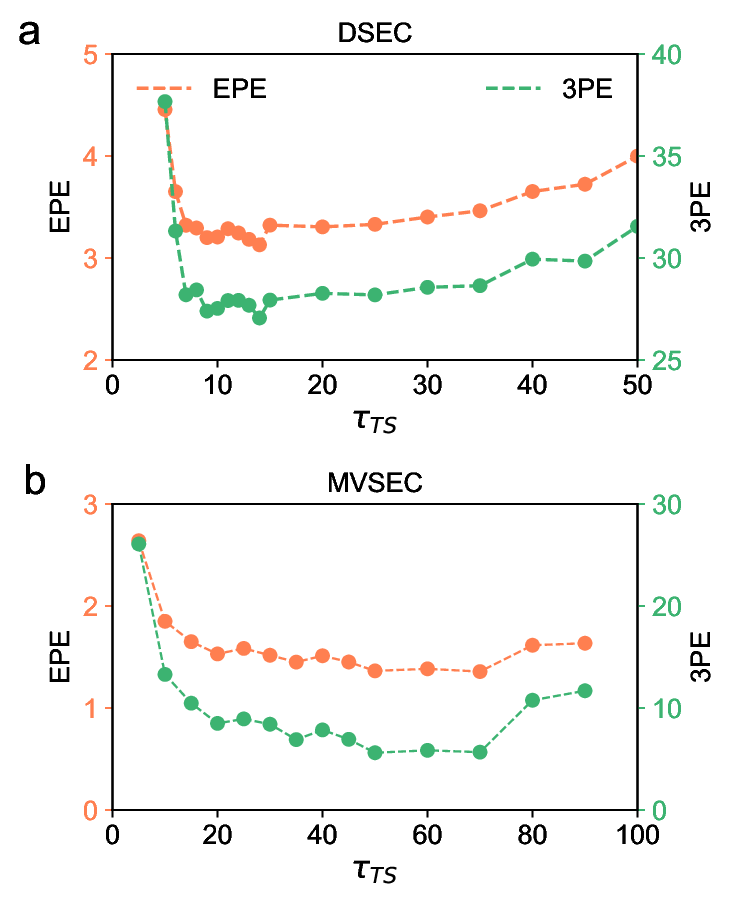} 
	\caption{\textbf{Effects of exponential-decay rate of time surface for VSA-Flow on DESC and MVSEC (the indoor\_flying1 sequence)}. Because the deviation of all metrics is small when $d=1024$ (Table \ref{ImpactDandSonDSEC}), for the sake of simplicity, every evaluation result for each $\tau_{TS}$ come from a single set of randomly generated HD kernel.
	}
	
	\label{fig:Effect_TauTS}
\end{figure}

\subsection{Effects of Exponential-decay Rate of Time Surface}
The temporal information of the HD feature descriptor is primarily impacted by the exponential-decay rate $\tau _{TS}$ of the accumulative Time Surface (TS). Figure \ref{fig:Effect_TauTS} illustrates the metrics EPE and 3PE for a single trial of the VSA-Flow method on DSEC and MVSEC datasets with varying $\tau_{TS}$. Both metrics exhibit a trend of initially decreasing and then increasing with $\tau_{TS}$. These results indicate that the performance of VSA-Flow's optical flow estimation diminishes when $\tau_{TS}$ is either too small or too large. Optimal performance is observed within a suitable range of $\tau_{TS}$. It is because short $\tau _{TS}$ for TS emphasizes recent events, resulting in its sparsity and inadequacy. Conversely, an excessively long $\tau _{TS}$ causes TS to encompass events over an extended period, leading to a blurred representation. Hence, an appropriate $\tau _{TS}$ is essential. It is worthy noting that the optimal range for $\tau_{TS}$ in the VSA-Flow method differs between DSEC and MVSEC due to variations in the characteristics of the event cameras used. In contrast to DSEC, events in MVSEC are more sparse, requiring a larger $\tau_{TS}$. This indicates the necessity to accumulate events over a longer period for MVSEC to achieve more accurate information encoding in the TS.

\section{Conclusions and Discussions}
\noindent In summary, our work introduces a novel VSA-based feature matching framework for event-based optical flow, applicable to both model-based (VSA-Flow) and self-supervised learning (VSA-SM) methods. The key of our work lies in the effective utilization of a VSA-based HD feature descriptor for event frames. The proposed methods can achieve accurate estimation of event-based optical flow in the feature matching methodology without restoring luminance or additional sensor information \cite{zhu2018ev, hagenaars2021self, deng2021learning, ding2022spatio, wan2022learning}. This work signifies an important advancement in event-based optical flow within the feature matching methodology, underscored by our compelling and robust results. The proposed framework can have broad applicability, extending to more event-based tasks such as depth estimation and tracking. 

Currently, most primary methods for event-based optical flow estimation applicable to both model-based and self-supervised learning are contrast maximization methods \cite{shiba2022secrets, ye2020unsupervised, paredes2021back, hagenaars2021self,  paredes2023taming}. The contrast maximization (CM) methods excel in utilizing temporal information from events but are less adept for local spatial features from events. Hence, these methods perform well in estimating optical flow within short time intervals or for small flow magnitudes. They require more complex strategies for achieving satisfactory performance in larger time intervals, such as producing sharp image of warped events (IWE) at multiple reference times through iterative warping \cite{hagenaars2021self, paredes2023taming}. In contrast, our methods, based on feature similarity maximization, excel in utilizing the local spatial features of events, but is comparatively weaker in exploiting temporal information. Consequently, our methods demonstrate better performance in optical estimation within larger time intervals (Table \ref{ResultsMVSEC1}). Our methods achieve competitive performance without complex strategies, and avoid circumvent issues such as occlusions and overfitting observed when warping events in CM methods \cite{shiba2022secrets}. Future research will focus on enhancing the temporal encoding capability in HD feature descriptors.

Traditionally, the feature matching is primarily determined by the differences between two local images within the neighborhoods of two feature points, which are often quantified using metrics such as the sum of absolute differences and Euclidean distance \cite{lagorce2016hots, liu2018adaptive, zhou2021event}. This approach is frequently applied in event camera hardware platforms \cite{liu2022edflow}. However, due to the inherent randomness in events, it may not be the most effective approach for gauging feature similarity directly from local event frames. Inspired from \cite{frady2021computing, renner2022neuromorphicVO}, we utilize the VSA-based HD kernel to extract the local features and structured symbolic representation to achieve feature fusion from both event polarities and multiple spatial scales. These approaches enhance the similarity of flow-matching feature descriptors as shown in our evaluation results. VSA, also known as Hyperdimensional Computing, is considered a new emerging neuromorphic computing model for ultra-efficient edge AI \cite{karunaratne2020memory, amrouch2022brain, zou2022eventhd}. Presently, our method focuses on dense optical flow estimation. With appropriate adjustments and configurations, our method is promising to efficiently and rapidly achieve sparse optical flow estimation on hardware, facilitating the design of event-driven hardware optical flow sensors \cite{chao2013comparative, honegger2013open, liu2022edflow}.

\backmatter

\bmhead{Acknowledgements}

\section*{Data Availability Statement}
The DSEC-Flow dataset is available for download from the website at  \href{https://dsec.ifi.uzh.ch/dsec-datasets/download/}{https://dsec.ifi.uzh.ch/dsec-datasets/download/}. 

\noindent The MVSEC dataset is available for download from the website at \href{https://daniilidis-group.github.io/mvsec/download/}{https://daniilidis-group.github.io/mvsec/download/}.

\bigskip

\begin{appendices}

\section{Parameter Configurations}

\begin{table*}[h]
	\centering
	\caption{Parameter Configurations}
	\small 
	\setlength{\tabcolsep}{1mm}
	\label{Parameters}
	\begin{tabular}{llcccc}
		\toprule
		\multirow{2}{10mm}{Parameters} & \multicolumn{1}{l}{\multirow{2}{35mm}{}} & \multicolumn{2}{c}{VSA-Flow} & \multicolumn{2}{c}{VSA-SM}  \\
		\cmidrule{3-6}          & \multicolumn{1}{l}{} & DSEC   & MVSEC   & DSEC   & MVSEC   \\
		
		\midrule
		\multicolumn{6}{l}{Accumulative time surface}  \\
		\midrule
		
		$\tau _{TS}$ & The exponential-decay rate & 35$ms$$^*$ & 35$ms$$^*$ & 35$ms$ & 35$ms$ \\
		
		\midrule
		\multicolumn{6}{l}{VSA-based kernel and HD feature descriptor}  \\
		\midrule
		
		$d$ & The hypervector dimension & 1024$^*$ & 1024 & 1024$^*$ & 1024 \\
		$N$ & The size of HD kernel & 21 & 25 & 21 & 25 \\
		$\sigma_K$ & Stander deviation of Gaussian kernel $G$ & 1.5 & 1.5 & 1.5 & 1.5 \\
		$S$ & The multi-scale number in the descriptor & 2$^*$ & 2 & 2$^*$ & 2 \\
		
		\midrule
		\multicolumn{6}{l}{The cost volume module and the optical flow estimator in VSA-Flow}  \\
		\midrule
		
		$M$ & The neighborhood size of the cost volume & 31 & 31 &   &   \\
		$\alpha$ & A coefficient in the optical flow probability volume  & 0.85 & 0.60 &   &   \\
		$s_c$ & The kernel size of average pooling  & 71 & 71 &   &   \\
		
		\midrule
		\multicolumn{6}{l}{Loss functions in VSA-SM}  \\
		\midrule
		
		$\lambda$ & The weight of the smoothness in loss function  &   &   & 1.0 &  1.0 \\
		$\alpha$ & A coefficient in $\mathcal{L} _{similarity}$  &   &   & 5.0  & 5.0  \\
		
		
		%
		%

		\bottomrule
	\end{tabular}%
	
	\begin{threeparttable}
		\begin{tablenotes}
			\item $^*$Unless explicitly noted. 
		\end{tablenotes}
	\end{threeparttable}
\end{table*}%

Table \ref{Parameters} shows parameter values used in proposed VSA-Flow and VSA-SM methods.

\end{appendices}


\bigskip
\newpage

\bibliographystyle{sn-basic}
\bibliography{sn-bibliography}

\begin{thebibliography}{66}
\providecommand{\natexlab}[1]{#1}
\providecommand{\url}[1]{{#1}}
\providecommand{\urlprefix}{URL }
\providecommand{\doi}[1]{\url{https://doi.org/#1}}
\providecommand{\eprint}[2][]{\url{#2}}
 \bibcommenthead

\bibitem[{Akolkar et~al(2020)Akolkar, Ieng, and Benosman}]{akolkar2020real}
Akolkar H, Ieng SH, Benosman R (2020) Real-time high speed motion prediction
  using fast aperture-robust event-driven visual flow. IEEE Transactions on
  Pattern Analysis and Machine Intelligence 44(1):361--372

\bibitem[{Almatrafi and Hirakawa(2019)}]{almatrafi2019davis}
Almatrafi M, Hirakawa K (2019) Davis camera optical flow. IEEE Transactions on
  Computational Imaging 6:396--407

\bibitem[{Almatrafi et~al(2020)Almatrafi, Baldwin, Aizawa, and
  Hirakawa}]{almatrafi2020distance}
Almatrafi M, Baldwin R, Aizawa K, et~al (2020) Distance surface for event-based
  optical flow. IEEE transactions on pattern analysis and machine intelligence
  42(7):1547--1556

\bibitem[{Amrouch et~al(2022)Amrouch, Imani, Jiao, Aloimonos, Fermuller, Yuan,
  Ma, Barkam, Genssler, and Sutor}]{amrouch2022brain}
Amrouch H, Imani M, Jiao X, et~al (2022) Brain-inspired hyperdimensional
  computing for ultra-efficient edge ai. In: 2022 International Conference on
  Hardware/Software Codesign and System Synthesis (CODES+ ISSS), IEEE, pp
  25--34

\bibitem[{Benosman et~al(2012)Benosman, Ieng, Clercq, Bartolozzi, and
  Srinivasan}]{benosman2012asynchronous}
Benosman R, Ieng SH, Clercq C, et~al (2012) Asynchronous frameless event-based
  optical flow. Neural Networks 27:32--37

\bibitem[{Benosman et~al(2013)Benosman, Clercq, Lagorce, Ieng, and
  Bartolozzi}]{benosman2013event}
Benosman R, Clercq C, Lagorce X, et~al (2013) Event-based visual flow. IEEE
  transactions on neural networks and learning systems 25(2):407--417

\bibitem[{Black and Anandan(1996)}]{black1996robust}
Black MJ, Anandan P (1996) The robust estimation of multiple motions:
  Parametric and piecewise-smooth flow fields. Computer vision and image
  understanding 63(1):75--104

\bibitem[{Brebion et~al(2022)Brebion, Moreau, and Davoine}]{brebion2022real}
Brebion V, Moreau J, Davoine F (2022) Real-time optical flow for vehicular
  perception with low-and high-resolution event cameras. IEEE Transactions on
  Intelligent Transportation Systems 23(9)

\bibitem[{Cao et~al(2023)Cao, Zhang, Luo, Peng, Lin, Yang, and
  Li}]{cao2023learning}
Cao YJ, Zhang XS, Luo FY, et~al (2023) Learning generalized visual odometry
  using position-aware optical flow and geometric bundle adjustment. Pattern
  Recognition 136:109262

\bibitem[{Chao et~al(2013)Chao, Gu, Gross, Guo, Fravolini, and
  Napolitano}]{chao2013comparative}
Chao H, Gu Y, Gross J, et~al (2013) A comparative study of optical flow and
  traditional sensors in uav navigation. In: 2013 American Control Conference,
  IEEE, pp 3858--3863

\bibitem[{Deng et~al(2021)Deng, Chen, Chen, and Li}]{deng2021learning}
Deng Y, Chen H, Chen H, et~al (2021) Learning from images: A distillation
  learning framework for event cameras. IEEE Transactions on Image Processing
  30:4919--4931

\bibitem[{Dewulf et~al(2023)Dewulf, De~Baets, and
  Stock}]{dewulf2023hyperdimensional}
Dewulf P, De~Baets B, Stock M (2023) The hyperdimensional transform for
  distributional modelling, regression and classification. arXiv preprint
  arXiv:231108150

\bibitem[{Ding et~al(2022)Ding, Zhao, Zhang, Gao, Xiong, Yu, and
  Huang}]{ding2022spatio}
Ding Z, Zhao R, Zhang J, et~al (2022) Spatio-temporal recurrent networks for
  event-based optical flow estimation. In: Proceedings of the AAAI conference
  on artificial intelligence, pp 525--533

\bibitem[{Frady et~al(2021)Frady, Kleyko, Kymn, Olshausen, and
  Sommer}]{frady2021computing}
Frady EP, Kleyko D, Kymn CJ, et~al (2021) Computing on functions using
  randomized vector representations. arXiv preprint arXiv:210903429

\bibitem[{Frady et~al(2022)Frady, Kleyko, Kymn, Olshausen, and
  Sommer}]{frady2022computing}
Frady EP, Kleyko D, Kymn CJ, et~al (2022) Computing on functions using
  randomized vector representations (in brief). In: Proceedings of the 2022
  Annual Neuro-Inspired Computational Elements Conference, pp 115--122

\bibitem[{Gallego et~al(2018)Gallego, Rebecq, and
  Scaramuzza}]{gallego2018unifying}
Gallego G, Rebecq H, Scaramuzza D (2018) A unifying contrast maximization
  framework for event cameras, with applications to motion, depth, and optical
  flow estimation. In: Proceedings of the IEEE conference on computer vision
  and pattern recognition, pp 3867--3876

\bibitem[{Gallego et~al(2019)Gallego, Gehrig, and
  Scaramuzza}]{gallego2019focus}
Gallego G, Gehrig M, Scaramuzza D (2019) Focus is all you need: Loss functions
  for event-based vision. In: Proceedings of the IEEE/CVF Conference on
  Computer Vision and Pattern Recognition, pp 12280--12289

\bibitem[{Gallego et~al(2020)Gallego, Delbr{\"u}ck, Orchard, Bartolozzi, Taba,
  Censi, Leutenegger, Davison, Conradt, Daniilidis et~al}]{gallego2020event}
Gallego G, Delbr{\"u}ck T, Orchard G, et~al (2020) Event-based vision: A
  survey. IEEE transactions on pattern analysis and machine intelligence
  44(1):154--180

\bibitem[{Ganesan et~al(2021)Ganesan, Gao, Gandhi, Raff, Oates, Holt, and
  McLean}]{ganesan2021learning}
Ganesan A, Gao H, Gandhi S, et~al (2021) Learning with holographic reduced
  representations. Advances in Neural Information Processing Systems
  34:25606--25620

\bibitem[{Gehrig et~al(2021{\natexlab{a}})Gehrig, Aarents, Gehrig, and
  Scaramuzza}]{gehrig2021dsec}
Gehrig M, Aarents W, Gehrig D, et~al (2021{\natexlab{a}}) Dsec: A stereo event
  camera dataset for driving scenarios. IEEE Robotics and Automation Letters
  6(3):4947--4954

\bibitem[{Gehrig et~al(2021{\natexlab{b}})Gehrig, Millh{\"a}usler, Gehrig, and
  Scaramuzza}]{gehrig2021raft}
Gehrig M, Millh{\"a}usler M, Gehrig D, et~al (2021{\natexlab{b}}) E-raft: Dense
  optical flow from event cameras. In: 2021 International Conference on 3D
  Vision (3DV), IEEE, pp 197--206

\bibitem[{Hagenaars et~al(2021)Hagenaars, Paredes-Vall{\'e}s, and
  De~Croon}]{hagenaars2021self}
Hagenaars J, Paredes-Vall{\'e}s F, De~Croon G (2021) Self-supervised learning
  of event-based optical flow with spiking neural networks. Advances in Neural
  Information Processing Systems 34:7167--7179

\bibitem[{Hersche et~al(2022)Hersche, Karunaratne, Cherubini, Benini,
  Sebastian, and Rahimi}]{hersche2022constrained}
Hersche M, Karunaratne G, Cherubini G, et~al (2022) Constrained few-shot
  class-incremental learning. In: Proceedings of the IEEE/CVF Conference on
  Computer Vision and Pattern Recognition, pp 9057--9067

\bibitem[{Hersche et~al(2023)Hersche, Zeqiri, Benini, Sebastian, and
  Rahimi}]{hersche2023neuro}
Hersche M, Zeqiri M, Benini L, et~al (2023) A neuro-vector-symbolic
  architecture for solving raven’s progressive matrices. Nature Machine
  Intelligence 5(4):363--375

\bibitem[{Honegger et~al(2013)Honegger, Meier, Tanskanen, and
  Pollefeys}]{honegger2013open}
Honegger D, Meier L, Tanskanen P, et~al (2013) An open source and open hardware
  embedded metric optical flow cmos camera for indoor and outdoor applications.
  In: 2013 IEEE International Conference on Robotics and Automation, IEEE, pp
  1736--1741

\bibitem[{Horn and Schunck(1981)}]{horn1981determining}
Horn BK, Schunck BG (1981) Determining optical flow. Artificial intelligence
  17(1-3):185--203

\bibitem[{Kanerva(2009)}]{kanerva2009hyperdimensional}
Kanerva P (2009) Hyperdimensional computing: An introduction to computing in
  distributed representation with high-dimensional random vectors. Cognitive
  computation 1:139--159

\bibitem[{Karunaratne et~al(2020)Karunaratne, Le~Gallo, Cherubini, Benini,
  Rahimi, and Sebastian}]{karunaratne2020memory}
Karunaratne G, Le~Gallo M, Cherubini G, et~al (2020) In-memory hyperdimensional
  computing. Nature Electronics 3(6):327--337

\bibitem[{Karunaratne et~al(2021)Karunaratne, Schmuck, Le~Gallo, Cherubini,
  Benini, Sebastian, and Rahimi}]{karunaratne2021robust}
Karunaratne G, Schmuck M, Le~Gallo M, et~al (2021) Robust high-dimensional
  memory-augmented neural networks. Nature communications 12(1):2468

\bibitem[{Karunaratne et~al(2022)Karunaratne, Hersche, Langeneager, Cherubini,
  Le~Gallo, Egger, Brew, Choi, Ok, Silvestre et~al}]{karunaratne2022memory}
Karunaratne G, Hersche M, Langeneager J, et~al (2022) In-memory realization of
  in-situ few-shot continual learning with a dynamically evolving explicit
  memory. In: ESSCIRC 2022-IEEE 48th European Solid State Circuits Conference
  (ESSCIRC), IEEE, pp 105--108

\bibitem[{Kempitiya et~al(2022)Kempitiya, De~Silva, Kahawala, Haputhanthri,
  Alahakoon, and Osipov}]{kempitiya2022parameterization}
Kempitiya T, De~Silva D, Kahawala S, et~al (2022) Parameterization of vector
  symbolic approach for sequence encoding based visual place recognition. In:
  2022 International Joint Conference on Neural Networks (IJCNN), IEEE, pp 1--7

\bibitem[{Kingma and Ba(2014)}]{kingma2014adam}
Kingma DP, Ba J (2014) Adam: A method for stochastic optimization. arXiv
  preprint arXiv:14126980

\bibitem[{Kleyko et~al(2021)Kleyko, Rachkovskij, Osipov, and
  Rahimi}]{kleyko2021survey}
Kleyko D, Rachkovskij DA, Osipov E, et~al (2021) A survey on hyperdimensional
  computing aka vector symbolic architectures, part i: Models and data
  transformations. ACM Computing Surveys (CSUR)

\bibitem[{Kleyko et~al(2023)Kleyko, Rachkovskij, Osipov, and
  Rahimi}]{kleyko2023survey}
Kleyko D, Rachkovskij D, Osipov E, et~al (2023) A survey on hyperdimensional
  computing aka vector symbolic architectures, part ii: Applications, cognitive
  models, and challenges. ACM Computing Surveys 55(9):1--52

\bibitem[{Komer(2020)}]{komer2020biologically}
Komer B (2020) Biologically inspired spatial representation

\bibitem[{Lagorce et~al(2016)Lagorce, Orchard, Galluppi, Shi, and
  Benosman}]{lagorce2016hots}
Lagorce X, Orchard G, Galluppi F, et~al (2016) Hots: a hierarchy of event-based
  time-surfaces for pattern recognition. IEEE transactions on pattern analysis
  and machine intelligence 39(7):1346--1359

\bibitem[{Lee et~al(2020)Lee, Kosta, Zhu, Chaney, Daniilidis, and
  Roy}]{lee2020spike}
Lee C, Kosta AK, Zhu AZ, et~al (2020) Spike-flownet: event-based optical flow
  estimation with energy-efficient hybrid neural networks. In: European
  Conference on Computer Vision, Springer, pp 366--382

\bibitem[{Li et~al(2023)Li, Huang, Chen, Shi, Li, Bao, Cui, and
  Zhang}]{li2023blinkflow}
Li Y, Huang Z, Chen S, et~al (2023) Blinkflow: A dataset to push the limits of
  event-based optical flow estimation. arXiv preprint arXiv:230307716

\bibitem[{Liu et~al(2023)Liu, Chen, Qu, Zhang, Li, Knoll, and
  Jiang}]{liu2023tma}
Liu H, Chen G, Qu S, et~al (2023) Tma: Temporal motion aggregation for
  event-based optical flow. arXiv preprint arXiv:230311629

\bibitem[{Liu and Delbruck(2018)}]{liu2018adaptive}
Liu M, Delbruck T (2018) Adaptive time-slice block-matching optical flow
  algorithm for dynamic vision sensors. BMVC

\bibitem[{Liu and Delbruck(2022)}]{liu2022edflow}
Liu M, Delbruck T (2022) Edflow: Event driven optical flow camera with keypoint
  detection and adaptive block matching. IEEE Transactions on Circuits and
  Systems for Video Technology 32(9):5776--5789

\bibitem[{Lucas and Kanade(1981)}]{lucas1981iterative}
Lucas BD, Kanade T (1981) An iterative image registration technique with an
  application to stereo vision. In: IJCAI'81: 7th international joint
  conference on Artificial intelligence, pp 674--679

\bibitem[{M{\'e}min and P{\'e}rez(2002)}]{memin2002hierarchical}
M{\'e}min E, P{\'e}rez P (2002) Hierarchical estimation and segmentation of
  dense motion fields. International Journal of Computer Vision 46:129--155

\bibitem[{Nagata et~al(2021)Nagata, Sekikawa, and Aoki}]{nagata2021optical}
Nagata J, Sekikawa Y, Aoki Y (2021) Optical flow estimation by matching time
  surface with event-based cameras. Sensors 21(4):1150

\bibitem[{Neubert and Schubert(2021)}]{neubert2021hyperdimensional}
Neubert P, Schubert S (2021) Hyperdimensional computing as a framework for
  systematic aggregation of image descriptors. In: Proceedings of the IEEE/CVF
  conference on computer vision and pattern recognition, pp 16938--16947

\bibitem[{Paredes-Vall{\'e}s and de~Croon(2021)}]{paredes2021back}
Paredes-Vall{\'e}s F, de~Croon GC (2021) Back to event basics: Self-supervised
  learning of image reconstruction for event cameras via photometric constancy.
  In: Proceedings of the IEEE/CVF Conference on Computer Vision and Pattern
  Recognition, pp 3446--3455

\bibitem[{Paredes-Vall{\'e}s et~al(2023)Paredes-Vall{\'e}s, Scheper, De~Wagter,
  and de~Croon}]{paredes2023taming}
Paredes-Vall{\'e}s F, Scheper KY, De~Wagter C, et~al (2023) Taming contrast
  maximization for learning sequential, low-latency, event-based optical flow.
  arXiv preprint arXiv:230305214

\bibitem[{Plate(1992)}]{plate1992holographic}
Plate TA (1992) Holographic recurrent networks. Advances in neural information
  processing systems 5

\bibitem[{Plate(1994)}]{plate1994distributed}
Plate TA (1994) Distributed representations and nested compositional structure.
  Citeseer

\bibitem[{Renner et~al(2022{\natexlab{a}})Renner, Supic, Danielescu, Indiveri,
  Frady, Sommer, and Sandamirskaya}]{renner2022neuromorphicVO}
Renner A, Supic L, Danielescu A, et~al (2022{\natexlab{a}}) Neuromorphic visual
  odometry with resonator networks. arXiv preprint arXiv:220902000

\bibitem[{Renner et~al(2022{\natexlab{b}})Renner, Supic, Danielescu, Indiveri,
  Olshausen, Sandamirskaya, Sommer, and Frady}]{renner2022neuromorphic}
Renner A, Supic L, Danielescu A, et~al (2022{\natexlab{b}}) Neuromorphic visual
  scene understanding with resonator networks. arXiv preprint arXiv:220812880

\bibitem[{Shiba et~al(2022)Shiba, Aoki, and Gallego}]{shiba2022secrets}
Shiba S, Aoki Y, Gallego G (2022) Secrets of event-based optical flow. In:
  European Conference on Computer Vision, Springer, pp 628--645

\bibitem[{Stoffregen and Kleeman(2018)}]{stoffregen2018simultaneous}
Stoffregen T, Kleeman L (2018) Simultaneous optical flow and segmentation
  (sofas) using dynamic vision sensor. arXiv preprint arXiv:180512326

\bibitem[{Stoffregen et~al(2020)Stoffregen, Scheerlinck, Scaramuzza, Drummond,
  Barnes, Kleeman, and Mahony}]{stoffregen2020reducing}
Stoffregen T, Scheerlinck C, Scaramuzza D, et~al (2020) Reducing the
  sim-to-real gap for event cameras. In: Computer Vision--ECCV 2020: 16th
  European Conference, Glasgow, UK, August 23--28, 2020, Proceedings, Part
  XXVII 16, Springer, pp 534--549

\bibitem[{Stone et~al(2021)Stone, Maurer, Ayvaci, Angelova, and
  Jonschkowski}]{stone2021smurf}
Stone A, Maurer D, Ayvaci A, et~al (2021) Smurf: Self-teaching multi-frame
  unsupervised raft with full-image warping. In: Proceedings of the IEEE/CVF
  conference on Computer Vision and Pattern Recognition, pp 3887--3896

\bibitem[{Sun et~al(2018)Sun, Yang, Liu, and Kautz}]{sun2018pwc}
Sun D, Yang X, Liu MY, et~al (2018) Pwc-net: Cnns for optical flow using
  pyramid, warping, and cost volume. In: Proceedings of the IEEE conference on
  computer vision and pattern recognition, pp 8934--8943

\bibitem[{Teed and Deng(2020)}]{teed2020raft}
Teed Z, Deng J (2020) Raft: Recurrent all-pairs field transforms for optical
  flow. In: Computer Vision--ECCV 2020: 16th European Conference, Glasgow, UK,
  August 23--28, 2020, Proceedings, Part II 16, Springer, pp 402--419

\bibitem[{Wan et~al(2022)Wan, Dai, and Mao}]{wan2022learning}
Wan Z, Dai Y, Mao Y (2022) Learning dense and continuous optical flow from an
  event camera. IEEE Transactions on Image Processing 31:7237--7251

\bibitem[{Wu et~al(2022)Wu, Paredes-Vall{\'e}s, and
  de~Croon}]{wu2022lightweight}
Wu Y, Paredes-Vall{\'e}s F, de~Croon GC (2022) Lightweight event-based optical
  flow estimation via iterative deblurring. arXiv preprint arXiv:221113726

\bibitem[{Ye et~al(2020)Ye, Mitrokhin, Ferm{\"u}ller, Yorke, and
  Aloimonos}]{ye2020unsupervised}
Ye C, Mitrokhin A, Ferm{\"u}ller C, et~al (2020) Unsupervised learning of dense
  optical flow, depth and egomotion with event-based sensors. In: 2020 IEEE/RSJ
  International Conference on Intelligent Robots and Systems (IROS), IEEE, pp
  5831--5838

\bibitem[{Ye et~al(2023)Ye, Shi, Yang, Wang, Yin, Wang, and
  Wang}]{ye2023towards}
Ye Y, Shi H, Yang K, et~al (2023) Towards anytime optical flow estimation with
  event cameras. arXiv preprint arXiv:230705033

\bibitem[{Zhang et~al(1988)Zhang, Tanida, Itoh, and Ichioka}]{zhang1988shift}
Zhang W, Tanida J, Itoh K, et~al (1988) Shift-invariant pattern recognition
  neural network and its optical architecture. In: Proceedings of annual
  conference of the Japan Society of Applied Physics, Montreal, CA

\bibitem[{Zhou et~al(2021)Zhou, Gallego, and Shen}]{zhou2021event}
Zhou Y, Gallego G, Shen S (2021) Event-based stereo visual odometry. IEEE
  Transactions on Robotics 37(5):1433--1450

\bibitem[{Zhu and Yuan(2018)}]{zhu2018ev}
Zhu AZ, Yuan L (2018) Ev-flownet: Self-supervised optical flow estimation for
  event-based cameras. In: Robotics: Science and Systems

\bibitem[{Zhu et~al(2019)Zhu, Yuan, Chaney, and
  Daniilidis}]{zhu2019unsupervised}
Zhu AZ, Yuan L, Chaney K, et~al (2019) Unsupervised event-based learning of
  optical flow, depth, and egomotion. In: Proceedings of the IEEE/CVF
  Conference on Computer Vision and Pattern Recognition, pp 989--997

\bibitem[{Zou et~al(2022)Zou, Alimohamadi, Kim, Najafi, Srinivasa, and
  Imani}]{zou2022eventhd}
Zou Z, Alimohamadi H, Kim Y, et~al (2022) Eventhd: Robust and efficient
  hyperdimensional learning with neuromorphic sensor. Frontiers in Neuroscience
  16:858329

\end{thebibliography}

\end{document}